# Frequency-based tension assessment of an inclined cable with complex boundary conditions using the PSO algorithm

Wen-ming Zhang[*,1], Zhi-wei Wang[a,1], Dan-dian Feng[b,1], Zhao Liu[c]

*The Key Laboratory of Concrete and Prestressed Concrete Structures of the Ministry of Education, Southeast University, Nanjing 211189, China*

**Abstract**: The frequency-based method is the most commonly used method for measuring cable tension. However, the calculation formulas for the conventional frequency-based method are generally based on the ideally hinged or fixed boundary conditions without a comprehensive consideration of the inclination angle, sag-extensibility, and flexural stiffness of cables, leading to a significant error in cable tension identification. This study aimed to propose a frequency-based method of cable tension identification considering the complex boundary conditions at the two ends of cables using the particle swarm optimization (PSO) algorithm. First, the refined stay cable model was established considering the inclination angle, flexural stiffness, and sag-extensibility, as well as the rotational constraint stiffness and lateral support stiffness for the unknown boundaries of cables. The vibration mode equation of the stay cable model was discretized and solved using the finite difference method. Then, a multiparameter identification method based on the PSO algorithm was proposed. This method was able to identify the tension, flexural stiffness, axial stiffness, boundary rotational constraint stiffness, and boundary lateral support stiffness according to the measured multiorder frequencies in a synchronous manner. The feasibility and accuracy of this method were validated through numerical cases. Finally, the proposed approach was applied to the tension identification of the anchor span strands of a suspension bridge (Jindong Bridge) in China. The results of cable tension identification using the proposed method and the existing methods discussed in previous studies were compared with the on-site pressure ring measurement results. The comparison showed that the proposed approach had a high accuracy in cable tension identification. Moreover, the synchronous identification of the flexural stiffness, axial stiffness, boundary rotational constraint stiffness, and boundary lateral support stiffness was highly beneficial in improving the results of cable tension identification.

## 1. Introduction

Cable is one of the indispensable components of modern bridge engineering, such as stay cables in cable-stayed bridges, flexible hanger rods in half-through or through arch bridges, and main cables and hanger rods in suspension bridges. The accurate measurement of cable tension is an important guarantee for construction control and health monitoring of bridges. In engineering practice, the methods used for measuring cable tension include the lifting jack with pressure gauge method, pressure sensor method, magnetic flux method, fiber Bragg grating method, and frequency-based method. Among these, the frequency-based method is the primary method used for detecting cable tension during construction and operation periods because of its nondestructive nature, convenience, and high efficiency (Chen *et al.* 2018,

---

[*] Corresponding author, Ph.D., Associate Professor
  E-mail: zwm@seu.edu.cn
[a] Ph.D. Student
[b] Master Student
[c] Professor
[1] These authors contributed equally to this work and should be considered co-first authors.

Jeong *et al.* 2020, Kim *et al.* 2020). However, the length, inclination angle, axial stiffness, flexural stiffness, and boundary conditions vary greatly depending on the type of cable used. The conventional frequency-based method can hardly achieve high accuracy of cable tension identification for all kinds of cables because the aforementioned parameters are not taken into account comprehensively. Noteworthy is that the environmental temperature variation will also affect the cable tension identification results (Ma *et al.* 2021). The frequency-based method discussed here focuses on cable tension identification under a certain temperature. Different modification schemes have been proposed based on the following two primary aspects to improve the accuracy of cable tension identification using the frequency-based method: flexural stiffness and sag-extensibility, and the influence of boundary conditions.

An unacceptable error may occur in cable tension identification if flexural stiffness and sag-extensibility are not considered (Casas 1994, Ren *et al.* 2005). The two most representative modification theories addressing this problem were proposed: the modern cable theory and the axially loaded beam theory. The modern cable theory fully considered the influence of only sag-extensibility of cables on frequency, and ignored the influence of flexural stiffness (Irvine 1981). Triantafyllou (1984) and Triantafyllou and Grinfogel (1986) believed that inclined cables and horizontal cables had different dynamic features. They proposed asymptotic equations for the natural frequencies and mode shapes of inclined cables. Russell and Lardner (1998) performed vibration experiments through which the simulation accuracy of the aforementioned approximate equations for vibration frequency of the stay cables was verified. However, they did not propose a specific method for cable tension identification. Therefore, some studies generally used antisymmetric modal frequencies or higher-order frequencies to achieve cable tension identification to eliminate the influence of sag-extensibility on cable tension identification (Fang and Wang 2012). Contrary to the modern cable theory, the axially loaded beam theory considered the influence of flexural stiffness but not that of sag-extensibility. If the boundary of cables is hinged, the formula for the axially loaded beam theory is explicit and the cable tension and flexural stiffness can be simultaneously identified using the linear regression method based on multiorder frequency measurement. For cables with a consolidated boundary condition or elastic embedding, the cable tension is calculated through the iterative solution of transcendental equations. Moreover, when using low-order frequency to identify the tension of the cable with large sag-extensibility, the axially loaded beam theory produces a considerable error compared with the actual situation. The reason is mainly that large sag-extensibility of cables has a significant impact on lower-order frequencies.

Given the limitations of the aforementioned two theories, many studies attempted to establish a dynamic model of cables that took both the sag-extensibility and flexural stiffness into consideration, or described a practical formula for cable tension that considered both these parameters. Ni *et al.* (2002) treated large-diameter sagged cables as the combination of ideal cable element and virtual beam element. They established a finite element model of stay cables considering the sag-extensibility and flexural stiffness, which was applied to the vibration simulation of the side-span main cable of the Tsing Ma Bridge. The accuracy of the model was verified by comparing calculated frequencies against measured frequencies. Ricciardi and Saitta (2008) established a continuous model considering the flexural stiffness and sag-extensibility, which was used to calculate in-plane frequencies and mode shapes of cables. However, neither Ni *et al.* (2002) nor Ricciardi and Saitta (2008) provided the specific method for cable tension identification. Zui *et al.* (1996) represented the physical properties of cables as nondimensional parameters. They provided the practical formulas for cable tension considering the flexural stiffness and sag-extensibility of cables in the form of a piecewise function. The formulas are applicable for various cables with different sag-extensibility and flexural stiffness as far as the vibration of first- or second-order mode is measurable. Mehrabi and Tabatabai (1998) introduced a finite difference method for discretizing the vibration mode equation of the horizontal cable. This model considered the influence of the sag-extensibility and flexural stiffness on cable vibration. Ren *et al.* (2005) applied the energy method to solve the vibration equation of cables. They presented two groups of empirical formulas for solving the cable tension based on fundamental frequencies of cables, which considered the influence of sag-extensibility and flexural stiffness, respectively, through curve fitting. The aforementioned studies discussed only two special situations, namely, consolidated and hinged boundary conditions at the two ends of the cable. They did not adequately consider the influence of general boundary conditions on the dynamic features of cables.

As to the influence of boundary conditions, differences in boundary constraints change vibration frequencies of the cable system, leading to a nonnegligible error in cable tension identification. In engineering practice, boundary conditions of cables are usually not ideally fixed or hinged. For example, one end of the anchor span strand of the suspension bridge is connected to the cantilever linkage in the front anchor facet, while the other end is stacked with other strands in the splay saddle groove. The support at the two ends of cables vibrate along with the vibration of cables. Therefore, boundary conditions at the two ends of the cable belonged to the elastic support and elastic rotation. Xie and Li (2014) and Ma (2017) established the cable model considering the rotational constraint stiffness for boundaries. However, the cable model built by Xie and Li (2014) targeted the vertical hangers, for which the influence of sag-extensibility and inclination angle was not considered. The cable model built by Ma (2017) did not consider the lateral support stiffness for boundaries of cables. Xu *et al.* (2019) established a tri-segment suspender dynamic model according to the characteristics of the anchorage ends of the suspender. Nevertheless, the influence of inclination angle was not considered both in model and verification test. Yan *et al.* (2019) proposed a cable tension identification method based on the mode to eliminate the influence of complex boundary conditions of cables. They built the equivalent segment model of cables between the two points of the zero-vibration mode. Based on the equivalent segment model, the formula of the axially loaded beam theory with hinged boundary conditions was used to calculate the equivalent cable tension. However, this method had a heavy reliance on the accuracy of the vibration mode identification and did not consider the sag-extensibility and inclination angle of the cable. The aforementioned studies considered the complex boundary conditions of cables to a certain extent, while they did not incorporate a systematic consideration of the cable inclination angle, flexural stiffness, sag-extensibility, boundary rotational constraint stiffness, and boundary lateral support stiffness.

Besides the refinement of vibration models of cables, developing a parameter identification algorithm for the cable system with high accuracy and efficiency is another important way to improve the identification accuracy of cable tension. Liao *et al.* (2012) adopted the least-squares optimization to eliminate the error in frequencies calculated by the precise finite element model compared with measured frequencies. This method was able to determine the cable tension and other system parameters in a synchronous manner. Li *et al.* (2014) described an extended Kalman filter algorithm that could identify temporal variations in cable tension. Kim and Park (2007) established a finite element model considering the inclination angle, flexural stiffness, and sag-extensibility of cables. They also described a frequency-based sensitivity-updating (FBSU) algorithm based on the Newton-Raphson method. This method used measured multiorder frequencies for synchronous identification of multiple parameters of the cable system, including the cable tension. However, this model did not consider the influence of complex boundary conditions of cables. During the study on the use of a frequency-based method to identify the tension of stay cables, Ma (2017) established the finite difference method for stay cables considering the rotational constraint stiffness at the two ends of the cable. The FBSU algorithm was used for parameter identification of this model, including parameters of cable tension, flexural stiffness, axial stiffness, and boundary rotational stiffness. However, the algorithm was not verified through engineering cases. Furthermore, the Newton-Raphson method had a heavy reliance on the choice of initial values for parameter identification, and iterative calculation could hardly achieve a convergence. For example, the value range was usually large for the boundary stiffness of cables, and initial values were hard to determine. Therefore, this method had poor robustness in practical application.

In recent years, some studies attempted to introduce the heuristic algorithm into cable tension identification. Rango *et al.* (2019) and Zarbaf *et al.* (2018) used artificial neural network in the identification of cable tension, but their methods can only realize the single parameter (i.e. cable tension) identification in specific cable systems. Xie and Li (2014) applied the genetic algorithm (GA) to the inverse operation of a finite element model of arch bridge hanger rods. The tension, flexural stiffness, and rotational stiffness at the two ends of the hanger rod were identified according to measured multiorder frequencies. However, this model neither involved the sag-extensibility nor considered the lateral support stiffness for boundaries, thereby restricting its application. Dan *et al.* (2015) updated the finite element model of cable-damper system by particle swarm optimization (PSO) algorithm, and realized the identification of cable tension, moment of inertia and damping coefficient. Zarbaf *et al*. (2017) inheriting the model by Mehrabi and Tabatabai (1998), introduced the Newton-Raphson method for the simple cable model and applied PSO and GA algorithms to the finite difference cable model. They compared the results of cable tension identification using these three methods and found that the PSO algorithm was relatively better. However, they still failed to solve the potential problem of parameter identification errors caused by a lack of refined consideration of cable inclination angle and complex boundary conditions. Dan *et al.* (2018a) proposed a multistep and multiparameter identification method by analyzing the parameters identification of eight different complex cable systems via PSO algorithm, in which the cable system parameters can be identified step by step. However, the boundary conditions were used as known parameters rather than identification objects in their cable systems. Moreover, the boundaries of cables in experimental verification were only set as fixed rather than complex boundary conditions.

Taken together, the recent studies on cable tension identification using the frequency-based method generally had the following drawbacks: (1) The cable model did not simultaneously consider the flexural stiffness, axial stiffness, sag-extensibility, inclination angle, rotational constraint stiffness, and lateral support stiffness for boundaries. Neglecting any of the aforementioned parameters may influence the accuracy of cable tension identification. As a result, the cable model was applied only to some special cases, impairing the universality of the method. (2) Parameter identification algorithms used in these studies usually targeted only a few parameters to be identified. However, the algorithm convergence and identification results heavily relied on the choice of initial values when identifying multiple parameters simultaneously. The initial values of some parameters were hard to choose, such as rotational constraint stiffness and lateral support stiffness for boundaries. For this reason, the FBSU algorithm based on the Newton-Raphson method for parameter identification typically cannot converge in practice. An optimization algorithm with a weaker reliance on initial values is urgently required. (3) Parameter identification usually requires an equal amount of known conditions. However, this requirement is very rigorous for parameter identification considering multiple factors, and it is also hard to achieve in practice. A cable tension identification method based on a few known conditions is the most favored to ensure the universality of the method. Among the various optimization algorithms, the particle swarm optimization (PSO) algorithm has a smaller number of parameters to be configured. Moreover, it has a fast convergence speed and low computational complexity. Therefore, the PSO algorithm is considered a more efficient, intelligent algorithm.

To overcome the aforementioned defects, a refined cable model that simultaneously considered the cable inclination angle, sag-extensibility, flexural stiffness, boundary rotational stiffness, and lateral support stiffness was established and a cable tension identification approach with high adaptability was proposed by combining with the high-efficiency, stable PSO algorithm. This approach could identify the cable tension, flexural stiffness, axial stiffness, boundary rotational constraint stiffness, and boundary lateral support stiffness according to measured multiorder frequencies. First, the nonlinear motion equation and modal equation of the cable were derived, and the discretized model of the cable was built using the finite difference method. Next, the multiparameter identification method based on the PSO algorithm was proposed. Methodological validation was performed through a numerical case study of four cables with different properties. Finally, this approach was verified in the tension testing of anchor span strands in Jindong Bridge in China. The results of cable tension identification using the proposed method and the existing methods proposed in previous studies were compared with the on-site pressure ring measurement results. The comparison demonstrated the applicability and accuracy of the proposed method.

## 2. Free vibration analysis of a cable

### 2.1 Free vibration equation of a cable

The dynamic model of a stay cable and the coordinate system where this dynamic model is situated are shown in Fig. 1. The rectangular coordinate system has the point of the left support A as the origin. The direction AB and the direction normal to AB are taken as the $x$-coordinate and the $y$-coordinate, respectively. $H_A$ and $H_B$ are the chordwise components of the tension at the two ends of the cable. $H_m$ is the average of the chordwise component of cable tension; $L$ is the cable length; and $EA = EA(x)$ is the axial stiffness of the cable. This model thoroughly considered the influence of the inclination angle, flexural stiffness, and sag-extensibility of the cable.

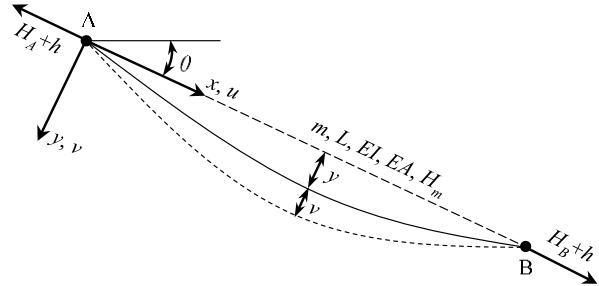

Fig. 1. A mathematical model for dynamic analysis of the stay cable.

Only the in-plane vibration of the cable is considered. It is assumed that no coupling occurred between the axial vibration and the transverse in-plane vibration of the cable. According to D'Alembert's principle, the following equations of in-plane chordwise and transverse free vibration of the cable are established:

$$\frac{\partial}{\partial s}\left[(T+\tau)\left(\frac{\mathrm{d}x}{\mathrm{d}s}+\frac{\partial u}{\partial s}\right)\right]+mg\sin\theta = m\frac{\partial^2 u}{\partial t^2}+c_u\frac{\partial u}{\partial t} \quad (1)$$

$$\frac{\partial}{\partial s}\left[(T+\tau)\left(\frac{\mathrm{d}y}{\mathrm{d}s}+\frac{\partial v}{\partial s}\right)\right]-\frac{\partial^2}{\partial s^2}\left(EI\frac{\partial^2 \eta}{\partial x^2}\right)+mg\cos\theta = m\frac{\partial^2 v}{\partial t^2}+c_v\frac{\partial v}{\partial t} \quad (2)$$

where $T = T(x)$ denotes the initial tensile force of the cable under the static equilibrium; $EI = EI(x)$ denotes the flexural stiffness of the cable at the position $x$; $\tau = \tau(x, t)$ denotes the additional tension caused by cable vibration; $s$ denotes the arc-length coordinate of the arc length of the cable; d$s$ denotes the tangential length of an infinitesimal cable element; $y = y(x)$ denotes the static line shape equation of the cable under the deadweight action; $u = u(x, t)$ and $v = v(x, t)$ are the displacement in the $x$ direction and $y$ direction caused by vibration, respectively; $\eta(x, t) = y + v$ denotes the sum of the cable displacement along the $y$ direction; $g$ denotes the gravitational acceleration; $m = m(x)$ denotes the mass of the cable per unit length; $\theta$ denotes the inclination angle of the cable; $c_u(x)$ and $c_v(x)$ represent the axial and transverse in-plane viscous damping coefficients per unit length, respectively.

Generally,

$$T + \tau = (H + h)\frac{\mathrm{d}s}{\mathrm{d}x} \tag{3}$$

where $H = H(x)$ is the chordwise component of $T(x)$, and $h = h(x, t)$ is the chordwise component of $\tau(x, t)$.

The cable tension increased gradually from the bottom to the top of the cable due to the influence of gravity and inclination angle. Thus,

$$H(x) = H_A - mgx\sin\theta \tag{4}$$

In addition, the axial vibration could be neglected. Therefore, Eq. (1) and Eq. (2) are, respectively, written as

$$\frac{\partial}{\partial s}(H + h) + mg\sin\theta = 0 \tag{5}$$

$$\frac{\partial}{\partial s}\left[(H + h)\left(\frac{\mathrm{d}y}{\mathrm{d}x} + \frac{\partial v}{\partial x}\right)\right] - \frac{\partial^2}{\partial s^2}\left(EI\frac{\partial^2 \eta}{\partial x^2}\right) + mg\cos\theta = m\frac{\partial^2 v}{\partial t^2} + c_v\frac{\partial v}{\partial t} \tag{6}$$

If the sag-to-span ratio of the cable is no more than 1/10, $\partial s$ can be replaced by $\partial x$. Then, according to Eq. (5), $\partial h/\partial x = 0$. Therefore, $h$ remains constant along the lengthwise direction of the cable, that is, $h(x, t) = h(t)$ is only a function of time.

The differential equation of static equilibrium of the stay cable is expressed as follows:

$$\frac{\mathrm{d}^2}{\mathrm{d}x^2}\left(EI\frac{\mathrm{d}^2 y}{\mathrm{d}x^2}\right) - H\frac{\mathrm{d}^2 y}{\mathrm{d}x^2} = mg\cos\theta \tag{7}$$

Substituting Eq. (7) into Eq. (6) and neglecting the second-order small quantities, the following expression is obtained after sorting:

$$H(x)\frac{\partial^2 v}{\partial x^2} + H'(x)\frac{\partial v}{\partial x} + h\frac{\mathrm{d}^2 y}{\mathrm{d}x^2} - \frac{\partial^2}{\partial x^2}\left(EI\frac{\partial^2 v}{\partial x^2}\right) = m\frac{\partial^2 v}{\partial t^2} + c_v\frac{\partial v}{\partial t} \tag{8}$$

Eq. (8) is the differential equation of the free vibration of the stay cable.

## 2.2 Vibration mode equation of a cable

Cable vibration can be considered as the superposition of several modes. The variable $v(x, t)$ is represented using the mode separation method by

$$v(x, t) = w(x)q(t) \tag{9}$$

where $w(x)$ is the mode shape function irrelevant to time $t$, and $q(t)$ is the generalized coordinate only relevant to time $t$. For the damped free vibration of the cable, $q(t)$ is usually represented by

$$q(t) = \mathrm{e}^{pt} \tag{10}$$

where $p$ is a complex number, given by

$$p = -\zeta\omega \pm \mathrm{i}\omega_D \tag{11}$$

where $\zeta$ is the damping ratio; i is the imaginary unit; $\omega$ is the circular frequency of the nondamping free vibration of the cable; and $\omega_D = \omega\sqrt{1 - \zeta^2}$ is the circular frequency of the damped free vibration of the cable.

After separating out the variable $v(x, t)$, Eq. (8) is written as

$$\frac{d^2}{dx^2}\left(EI\frac{d^2w}{dx^2}\right)q - H\frac{d^2w}{dx^2}q - H'\frac{dw}{dx}q - h\frac{d^2y}{dx^2} + c_v pwq + mp^2 wq = 0 \qquad (12)$$

If $h$ is represented in the form of linear expression of the generalized coordinate $q$, then the generalized coordinate $q$ can be extracted from Eq. (12) as the common factor. According to Hooke's law, the relationship between the additional cable tension ($\tau$) and the deformation relative to the initial static profile of a cable can be obtained. Irvine (1981) derived the following equation:

$$\frac{h}{EA}\left(\frac{ds}{dx}\right)^3 = \frac{\partial u}{\partial x} + \frac{dy}{dx} \times \frac{\partial v}{\partial x} \qquad (13)$$

Integrating Eq. (13) with respect to $(0, L)$ produces

$$\int_0^L \frac{h(ds/dx)^3}{EA} dx = \int_0^L \frac{\partial u}{\partial x} dx + \int_0^L \frac{dy}{dx} \frac{\partial v}{\partial x} dx \qquad (14)$$

As it is assumed that the chordwise displacement of the cable $u = 0$, the first term on the right side of Eq. (14) is zero. As $h$ is only a function of time, it is written as

$$h = \frac{\int_0^L \frac{dy}{dx} \frac{\partial v}{\partial x} dx}{\int_0^L \frac{(ds/dx)^3}{EA} dx} = \frac{\int_0^L \frac{dy}{dx} \frac{dw}{dx} dx}{\int_0^L \frac{(ds/dx)^3}{EA} dx} q = h_q q \qquad (15)$$

where $h_q$ is a constant irrelevant to time. Thus, $h$ is converted to the linear expression of the generalized coordinate $q$. Substituting Eq. (15) into Eq. (12) and eliminating the common factor $q$ produces

$$\frac{d^2}{dx^2}\left(EI\frac{d^2w}{dx^2}\right) - H\frac{d^2w}{dx^2} - H'\frac{dw}{dx} - h_q\frac{d^2y}{dx^2} + c_v pw + mp^2 w = 0 \qquad (16)$$

Expansion of the first term on the left side of Eq. (16) produces

$$\frac{d^2}{dx^2}\left(EI\frac{d^2w}{dx^2}\right) = \frac{d^2(EI)}{dx^2}\frac{d^2w}{dx^2} + 2\frac{d(EI)}{dx}\frac{d^3w}{dx^3} + EI\frac{d^4w}{dx^4} \qquad (17)$$

Replacing $ds$ in $h_q$ with $[(dx)^2 + (dy)^2]^{1/2}$, $h_q$ is represented as

$$h_q = \frac{\int_0^L \frac{dy}{dx} \frac{dw}{dx} dx}{\int_0^L \frac{\left[(dy/dx)^2 + 1\right]^{3/2}}{EA} dx} \qquad (18)$$

By integration by parts of the numerator, the following equation is obtained:

$$h_q = \frac{-\int_0^L \frac{d^2y}{dx^2} w \, dx}{\int_0^L \frac{\left[(dy/dx)^2 + 1\right]^{3/2}}{EA} dx} \qquad (19)$$

Substituting Eq. (17) and Eq. (19) into Eq. (16), the vibration mode equation of an inclined cable is derived as follows:

$$\frac{d^2(EI)}{dx^2}\frac{d^2w}{dx^2} + 2\frac{d(EI)}{dx}\frac{d^3w}{dx^3} + EI\frac{d^4w}{dx^4} - H\frac{d^2w}{dx^2} - H'\frac{dw}{dx} + \frac{\int_0^L \frac{d^2y}{dx^2} w \, dx}{\int_0^L \frac{\left[(dy/dx)^2 + 1\right]^{3/2}}{EA} dx} \frac{d^2y}{dx^2} + c_v pw + mp^2 w = 0 \qquad (20)$$

Next, the aforementioned fourth-order nonlinear differential equation was discretized and solved using the finite difference method.

## 2.3 Discretization of the mode equation using the finite difference method

The discretized model is shown in Fig. 2. The $n$ internal nodes are divided into $n + 1$ segments along the chordwise direction of the cable. The projected length of each cable segment along the chordwise direction of the cable is $a = L/(n + 1)$.

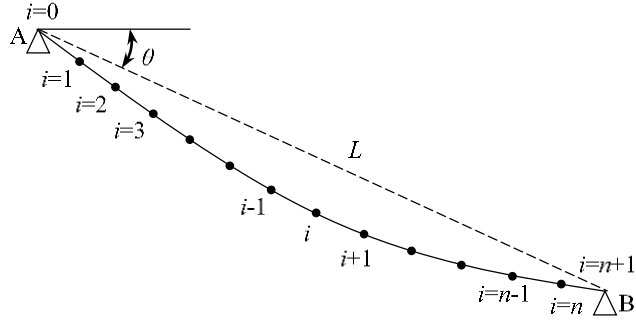

Fig. 2. Discretized model of the stay cable.

For the discretization of Eq. (20), taking the mode shape function $w(x)$ as an example, its differential format is defined as follows:

$$\frac{dw(x_i)}{dx} = \frac{w_{i+1} - w_{i-1}}{2a}, \quad \frac{d^2w(x_i)}{dx^2} = \frac{w_{i+1} - 2w_i + w_{i-1}}{a^2} \tag{21a, b}$$

$$\frac{d^3w(x_i)}{dx^3} = \frac{w_{i+2} - 2w_{i+1} + 2w_{i-1} - w_{i-2}}{2a^3}, \quad \frac{d^4w(x_i)}{dx^4} = \frac{w_{i+2} - 4w_{i+1} + 6w_i - 4w_{i-1} + w_{i-2}}{a^4} \tag{21c, d}$$

Furthermore, the differential formats of functions $EI(x)$, $EA(x)$, and $y(x)$ are similar to that of $w(x)$. The differential schemes of individual terms in Eq. (20) are transformed into different schemes after using the finite difference method for discretization. After discretization, the vibration mode equation in the matrix form is written as:

$$\boldsymbol{K}\boldsymbol{w} + \boldsymbol{C}p\boldsymbol{w} + \boldsymbol{M}p^2\boldsymbol{w} = 0 \tag{22}$$

where

$$\boldsymbol{K} = \boldsymbol{K}_1 + \boldsymbol{K}_2 \tag{23}$$

$$\boldsymbol{w}^T = \{w_1, w_2, \ldots, w_n\}; \boldsymbol{M} = \text{diag}\{m_1, m_2, \ldots, m_n\}; \boldsymbol{C} = \text{diag}\{c_{y1}, c_{y2}, \ldots, c_{yn}\} \tag{24a, b, c}$$

where $\boldsymbol{K}_1$ is the linear stiffness matrix of the cable corresponding to the first five terms on the left side of Eq. (20); and $\boldsymbol{K}_2$ is the nonlinear stiffness matrix of the cable corresponding to the sixth term on the left side of Eq. (20). $w_i$ and $m_i$ are the mode shape displacement and the mass per unit length at the internal node $i$ ($i = 1, 2, \ldots, n$), respectively. $c_{yi} = c_i/a$, $c_i$ is the viscous damping factor for damper connected to node $i$ ($i = 1, 2, \ldots, n$) along the $y$ direction.

The linear stiffness matrix $\boldsymbol{K}_1$ is represented as follows:

$$\boldsymbol{K}_1 = \begin{pmatrix} Q & U_1 & W_1 & & & & & 0 \\ R & S_2 & U_2 & W_2 & & & & \\ V_3 & D_3 & S_3 & U_3 & W_3 & & & \\ & - & - & - & - & & & \\ & & V_i & D_i & S_i & U_i & W_i & \\ & & & & V_{n-2} & S_{n-2} & U_{n-1} & P \\ 0 & & & & & V_n & D_n & G \end{pmatrix}_{n \times n} \tag{25}$$

where each row of the matrix has a one-to-one correspondence with each internal node of the discretized cable, where

$$V_i = -\frac{1}{2a^4}\left(EI_{i+1} - 2EI_i - EI_{i-1}\right) \tag{26a}$$

$$W_i = \frac{1}{2a^4}\left(EI_{i+1} + 2EI_i - EI_{i-1}\right) \tag{26b}$$

$$S_i = \frac{1}{a^4}\left(-2EI_{i+1} + 10EI_i - 2EI_{i-1}\right) + \frac{2H_i}{a^2} \tag{26c}$$

$$D_i = \frac{1}{a^4}\left(2EI_{i+1} - 6EI_i\right) - \frac{H_i}{a^2} + \frac{H_{i+1} - H_{i-1}}{4a^2} \tag{26d}$$

$$U_i = \frac{1}{a^4}\left(-6EI_i + 2EI_{i-1}\right) - \frac{H_i}{a^2} - \frac{H_{i+1} - H_{i-1}}{4a^2} \tag{26e}$$

in which $EI_i$ and $H_i$ are the flexural stiffness and the chordwise component of cable tension at the internal node $i$ ($i = 1, 2, …, n$), respectively. Compared with the previous studies on horizontal cables (Mebrabi and Tabatabai 1998), the influence of the inclination angle is incorporated into $D$ and $U$. Furthermore, the matrix elements $Q$, $G$, $R$, and $P$ are relevant to boundary conditions at the two ends of the cable. The equation of boundary equilibrium needs to be established at the two ends of the cable to solve these elements.

The nonlinear stiffness matrix $K_2$ is calculated using the following equation:

$$\underset{n \times n}{K_2} = rz^{\mathrm{T}} \tag{27}$$

where

$$r_i = \frac{z_i}{\sum_{i=1}^{n} t_i^3 / EA_i}; \quad z_i = \frac{y_{i+1} - 2y_i + y_{i-1}}{a^2}; \quad b_i = \left\{\left[\left(y_{i+1} - y_{i-1}\right)/2a\right]^2 + 1\right\}^{1/2} \tag{28a, b, c}$$

in which $y_i$ is the displacement of the internal node $i$ ($i = 1, 2, …, n$) under the static equilibrium along the $y$ direction.

The influence of damping in the cable system can be neglected in engineering practice because the damping ratio $\zeta$ of the cable structure is usually very small. In that case, $\sqrt{1-\zeta^2}$ approaches 1. That is to say, the frequency $\omega_D$ of damped free vibration of the cable is approximately equal to the circular frequency $\omega$ of the nondamping free vibration of the cable. Thus, $p^2 = -\omega^2$, and Eq. (22) is rewritten as

$$Kw - M\omega^2 w = 0 \tag{29}$$

According to Eq. (29), the modal frequency and mode shape of the free vibration of the cable are the generalized eigenvalue and eigenvector of the mass matrix and stiffness matrix of the cable, respectively. The mass matrix $M$ can be directly obtained from the material parameters or the actual situation. Therefore, as long as the intact linear stiffness matrix $K_1$ and nonlinear stiffness matrix $K_2$ are obtained, vibration frequencies of each order can be solved for the cable. Based on the aforementioned inferences, the matrix elements $Q$, $G$, $R$, and $P$ relevant to boundary conditions in $K_1$ are calculated through the analysis of boundary conditions of the cable. In addition, the static line shape of the cable also needs to be analyzed because some of the terms used depended on $y$, as reflected in the expression for the stiffness matrix $K_2$.

## 3. Equilibrium equations with boundary constraint

Boundary conditions at the two ends of the cable should be considered when solving the elements $Q$, $G$, $R$, and $P$ in the linear stiffness matrix $K_1$. According to Eq. (25) and Eq. (26), the product of $i$th row ($2 < i < n - 2$) of $K_1$ and the mode shape vector $w$ can be written as

$$V_i w_{i-2} + D_i w_{i-1} + S_i w_i + U_i w_{i+1} + W_i w_{i+2}$$
$$= \left[ -\frac{1}{2a^4} \left( EI_{i+1} - 2EI_i - EI_{i-1} \right) \right] w_{i-2} + \left[ \frac{1}{a^4} \left( 2EI_{i+1} - 6EI_i \right) - \frac{H_i}{a^2} + \frac{H_{i+1} - H_{i-1}}{4a^2} \right] w_{i-1}$$
$$+ \left[ \frac{1}{a^4} \left( -2EI_{i+1} + 10EI_i - 2EI_{i-1} \right) + \frac{2H_i}{a^2} \right] w_i + \left[ \frac{1}{a^4} \left( -6EI_i + 2EI_{i-1} \right) - \frac{H_i}{a^2} - \frac{H_{i+1} - H_{i-1}}{4a^2} \right] w_{i+1}$$
$$+ \left[ \frac{1}{2a^4} \left( EI_{i+1} + 2EI_i - EI_{i-1} \right) \right] w_{i+2} \quad (30)$$

Moreover, the product of the first row ($i = 1$) of the matrix $K_1$ and the mode shape vector $w$ should be equal to the right side of Eq. (30) ($i = 1$). Thus,

$$Qw_1 + U_1 w_2 + W_1 w_3$$
$$= \left[ -\frac{1}{2a^4} \left( EI_2 - 2EI_1 - EI_0 \right) \right] w_{-1} + \left[ \frac{1}{a^4} \left( 2EI_2 - 6EI_1 \right) - \frac{H_1}{a^2} + \frac{H_2 - H_0}{4a^2} \right] w_0$$
$$+ \left[ \frac{1}{a^4} \left( -2EI_2 + 10EI_1 - 2EI_0 \right) + \frac{2H_1}{a^2} \right] w_1 + \left[ \frac{1}{a^4} \left( -6EI_1 + 2EI_0 \right) - \frac{H_1}{a^2} - \frac{H_2 - H_0}{4a^2} \right] w_2$$
$$+ \left[ \frac{1}{2a^4} \left( EI_2 + 2EI_1 - EI_0 \right) \right] w_3 \quad (31)$$

In Eq. (31), there are two additional terms containing the factor $w_{-1}$ and $w_0$ on the right side compared with the left side. The one-to-one correspondence between the two sides of Eq. (31) can be established by replacing $w_{-1}$ and $w_0$ with a factor containing $w_1$ to determine the element $Q$. Similarly, according to the result of the last row of the matrix $K_1$ multiplied by the mode shape vector $w$, the one-to-one correspondence can be established by replacing $w_{n+1}$ and $w_{n+2}$ with a factor containing $w_n$ to determine the element $G$.

The product of the second row ($i = 2$) of the matrix $K_1$ and the mode shape vector $w$ should be equal to the right side of Eq. (30) ($i = 2$). Thus,

$$Rw_1 + S_2 w_2 + U_2 w_3 + W_2 w_4$$
$$= \left[ -\frac{1}{2a^4} \left( EI_3 - 2EI_2 - EI_1 \right) \right] w_0 + \left[ \frac{1}{a^4} \left( 2EI_3 - 6EI_2 \right) - \frac{H_2}{a^2} + \frac{H_3 - H_1}{4a^2} \right] w_1$$
$$+ \left[ \frac{1}{a^4} \left( -2EI_3 + 10EI_2 - 2EI_1 \right) + \frac{2H_2}{a^2} \right] w_2 + \left[ \frac{1}{a^4} \left( -6EI_2 + 2EI_1 \right) - \frac{H_2}{a^2} - \frac{H_3 - H_1}{4a^2} \right] w_3$$
$$+ \left[ \frac{1}{2a^4} \left( EI_3 + 2EI_2 - EI_1 \right) \right] w_4 \quad (32)$$

In Eq. (32) there is an additional term containing the factor $w_0$ on the right side compared with the left side. The one-to-one correspondence between the two sides of Eq. (32) can be established by replacing $w_0$ with a factor containing $w_1$ to determine the element $R$. Similarly, according to the product of the second-to-last row of the matrix $K_1$ with the mode shape vector $w$, the one-to-one correspondence can be established by replacing $w_{n+1}$ with a factor containing $w_n$ to determine the element $P$. Boundary constraint equilibrium equations of the cable are needed to achieve these goals.

In the present study, the refined vibration model of cables considering rotational constraint stiffness and lateral support stiffness at the two ends of the cable was established. The boundary constraint status of the cable is shown in Fig. 3, where $K_{r1}$, $K_{r2}$, $K_{s1}$, and $K_{s2}$ are the rotational constraint stiffness and lateral support stiffness at the left and right ends of the cable, respectively.

According to the equilibrium of bending moment and shear force at the ends of the cable, the following equations is derived:

$$K_{r1}\frac{dw}{dx}\bigg|_{x=0} - EI\frac{d^2w}{dx^2}\bigg|_{x=0} = 0, \quad K_{s1}w\big|_{x=0} - H\frac{dw}{dx}\bigg|_{x=0} + EI\frac{d^3w}{dx^3}\bigg|_{x=0} = 0 \qquad (33a, b)$$

$$K_{r2}\frac{dw}{dx}\bigg|_{x=l} + EI\frac{d^2w}{dx^2}\bigg|_{x=l} = 0, \quad K_{s2}w\big|_{x=l} + H\frac{dw}{dx}\bigg|_{x=l} - EI\frac{d^3w}{dx^3}\bigg|_{x=l} = 0 \qquad (33c, d)$$

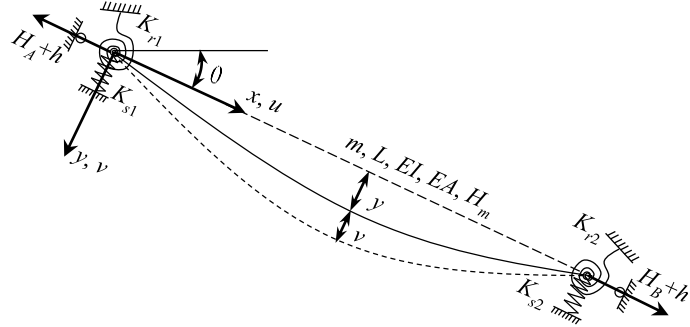

Fig. 3. An analytical model of the cable considering complex boundary conditions.

Discretization of the computational model of the cable produces

$$K_{r1}\frac{dw(x_0)}{dx} - EI_0\frac{d^2w(x_0)}{dx^2} = 0 \qquad (34a)$$

$$K_{s1}w(x_0) - H_0\frac{dw(x_0)}{dx} + \frac{d}{dx}\left(EI_0\frac{d^2w(x_0)}{dx^2}\right) = 0 \qquad (34b)$$

$$K_{r2}\frac{dw(x_{n+1})}{dx} + EI_{n+1}\frac{d^2w(x_{n+1})}{dx^2} = 0 \qquad (34c)$$

$$K_{s2}w(x_{n+1}) + H_{n+1}\frac{dw(x_{n+1})}{dx} - \frac{d}{dx}\left(EI_{n+1}\frac{d^2w(x_{n+1})}{dx^2}\right) = 0 \qquad (34d)$$

Substituting Eq. (34a) and Eq. (34c) into Eq. (34b) and Eq. (34d), respectively, produces

$$K_{s1}w(x_0) - H_0\frac{dw(x_0)}{dx} + K_{r1}\frac{d^2w(x_0)}{dx^2} = 0 \qquad (35a)$$

$$K_{s2}w(x_{n+1}) + H_{n+1}\frac{dw(x_{n+1})}{dx} + K_{r2}\frac{d^2w(x_{n+1})}{dx^2} = 0 \qquad (35b)$$

The differential forms of Eq. (34a), Eq. (34c), Eq. (35a), and Eq. (35b) are, respectively:

$$K_{r1}\frac{w_1 - w_{-1}}{2a} - EI_0\frac{w_1 - 2w_0 + w_{-1}}{a^2} = 0 \qquad (36a)$$

$$K_{s1}w_0 - H_0\frac{w_1 - w_{-1}}{2a} + K_{r1}\frac{w_1 - 2w_0 + w_{-1}}{a^2} = 0 \qquad (36b)$$

$$K_{r2}\frac{w_{n+2} - w_n}{2a} + EI_{n+1}\frac{w_{n+2} - 2w_{n+1} + w_n}{a^2} = 0 \qquad (36c)$$

$$K_{s2}w_{n+1} + H_{n+1}\frac{w_{n+2} - w_n}{2a} + K_{r2}\frac{w_{n+2} - 2w_{n+1} + w_n}{a^2} = 0 \qquad (36d)$$

A set of simultaneous equations is established using Eq. (36a), Eq. (36b), Eq. (36c), and Eq. (36d) and solved to obtain $w_0$, $w_{-1}$, $w_{n+1}$, and $w_{n+2}$ as follows:

$$w_0 = \frac{2EI_0H_0 - 2K_{r1}^2}{K_{s1}a(K_{r1}a + 2EI_0) + 2EI_0H_0 - 2K_{r1}^2}w_1 \qquad (37a)$$

$$w_{-1} = \frac{K_{r1}a - 2EI_0}{K_{r1}a + 2EI_0}w_1 + \frac{4EI_0}{K_{r1}a + 2EI_0} \times \frac{2EI_0H_0 - 2K_{r1}^2}{K_{s1}a(K_{r1}a + 2EI_0) + 2EI_0H_0 - 2K_{r1}^2}w_1 \tag{37b}$$

$$w_{n+1} = \frac{2EI_{n+1}H_{n+1} - 2K_{r2}^2}{K_{s2}a(K_{r2}a + 2EI_{n+1}) + 2EI_{n+1}H_{n+1} - 2K_{r2}^2}w_n \tag{37c}$$

$$w_{n+2} = \frac{(K_{r2}a - 2EI_{n+1})}{K_{r2}a + 2EI_{n+1}}w_n + \frac{4EI_{n+1}}{K_{r2}a + 2EI_{n+1}} \times \frac{2EI_{n+1}H_{n+1} - 2K_{r2}^2}{K_{s2}a(K_{r2}a + 2EI_{n+1}) + 2EI_{n+1}H_{n+1} - 2K_{r2}^2}w_n \tag{37d}$$

Substituting Eq. (37a) and Eq. (37b) into Eq. (31), the matrix element $Q$ is calculated as follows:

$$Q = S_1 + \frac{K_{r1}a - 2EI_0}{K_{r1}a + 2EI_0}V_1 + \frac{2EI_0H_0 - 2K_{r1}^2}{K_{s1}a(K_{r1}a + 2EI_0) + 2EI_0H_0 - 2K_{r1}^2}\left(D_1 + \frac{4EI_0}{K_{r1}a + 2EI_0}V_1\right) \tag{38a}$$

Substituting Eq. (37a) into Eq. (32), the matrix element $R$ is calculated as follows:

$$R = D_2 + \frac{2EI_0H_0 - 2K_{r1}^2}{K_{s1}a(K_{r1}a + 2EI_0) + 2EI_0H_0 - 2K_{r1}^2}V_2 \tag{38b}$$

Similarly, the matrix elements $G$ and $P$ are calculated as follows:

$$G = S_n + \frac{K_{r2}a - 2EI_{n+1}}{K_{r2}a + 2EI_{n+1}}W_n + \frac{2EI_{n+1}H_{n+1} - 2K_{r2}^2}{K_{s2}a(K_{r2}a + 2EI_{n+1}) + 2EI_{n+1}H_{n+1} - 2K_{r2}^2}\left(U_n + \frac{4EI_{n+1}}{K_{r2}a + 2EI_{n+1}}W_n\right) \tag{38c}$$

$$P = U_{n-1} + \frac{2EI_{n+1}H_{n+1} - 2K_{r2}^2}{K_{s2}a(K_{r2}a + 2EI_{n+1}) + 2EI_{n+1}H_{n+1} - 2K_{r2}^2}W_{n-1} \tag{38d}$$

where $EI_0$ and $EI_{n+1}$ are the flexural stiffness parameters at the two ends of the cable, respectively; $H_0$ and $H_{n+1}$ are the tension parameters at the two ends of the cable, respectively (i.e., $H_A$ and $H_B$ in Fig. 1). Based on the aforementioned inferences, Eq. (29) sufficiently considers the influence of complex boundary stiffness, and all elements of the linear stiffness matrix $\mathbf{K}_1$ were solved.

## 4. Static profile of a cable

As shown by the expression of each element in the stiffness matrix $K_2$, the static line shape of the cable is still needed to solve this finite difference model, that is, the displacement $y_i$ of each internal node $i$ ($i = 1, 2, …, n$) perpendicular to the chordwise direction under the static equilibrium condition. The static profile was usually approximated as parabolic (Irvine 1981, Kim and Park 2007, Ni *et al.* 2002). This assumption may be suitable for horizontal cables with a constant cross-sectional area and without considering the flexural stiffness. However, when the inclination angle and flexural stiffness of the cable cannot be ignored or the mass per unit length of the cable changes along the cable, the actual static line shape of the cable considerably deviates from the parabola. A precise static line shape enables a more accurate description of the influence of sag-extensibility on the cable system. In this study, the static line shape was determined by numerical calculation using the finite difference method. After discretization of the equation of static equilibrium, Eq. (7), it is also represented in the matrix form:

$$\mathbf{K}_s \mathbf{y} = \mathbf{m}g\cos\theta \tag{39}$$

$$\mathbf{y}^T = \{y_1, y_2, \cdots, y_n\}; \quad \mathbf{m}^T = \{m_1, m_2, \cdots, m_n\} \tag{40}$$

where $\mathbf{K}_s$ is the static stiffness matrix, whose form is similar to that of $\mathbf{K}_1$, although elements $D$ and $U$ need to be rewritten as:

$$D_i = \frac{1}{a^4}(2EI_{i+1} - 6EI_i) - \frac{H_i}{a^2} \tag{41a}$$

$$U_i = \frac{1}{a^4}(-6EI_i + 2EI_{i-1}) - \frac{H_i}{a^2} \tag{41b}$$

where each term is already defined earlier. Solving Eq. (39), the static line shape of the cable corresponding to the discretized model could be obtained.

## 5. Multiparameter identification via the PSO algorithm

The present study aimed to achieve cable tension identification based on on-site frequency measurement. This process involved other parameters of the cable system, such as flexural stiffness, axial stiffness, and boundary constraint stiffness. However, in engineering practice, these parameters are generally hard to estimate accurately in advance. The cross-sections of cables are usually made of several steel strands or parallel steel wires bound together, whose bending and shearing mechanisms may be different from those of the beams. Therefore, it may be unreliable to calculate the flexural stiffness $EI$ and axial stiffness $EA$ of the cable using the formulas designed for beams. In addition, boundary conditions at the two ends of cables in engineering practice are not ideally hinged or fixed, but rather the unknown state of elastic rotation and elastic support. To identify the cable tension $H$, it is also necessary to identify the flexural stiffness $EI$ and axial stiffness $EA$ of the cable, as well as the rotational constraint stiffness $K_{r1}$ and $K_{r2}$ and lateral support stiffness $K_{s1}$ and $K_{s2}$ at the two ends of the cable. The cable tension identified in this study was the average $H_m$ of its chordwise component. Conventional numerical algorithms, such as the Newton-Raphson method, have a heavy reliance on the choice of initial values as long as the convergence is concerned. Moreover, a large number of cable parameters needed to be identified in the present study, so it is hard to identify all the parameters synchronously by Newton-Raphson method. The PSO algorithm, as a swarm intelligent optimization algorithm, effectively reduced the influence of initial values on the algorithm convergence performance, thus achieving synchronous identification of several parameters. Compared with other intelligent optimization algorithms such as GA, fewer parameters needed to be configured with the PSO algorithm and the calculation efficiency was higher. In this study, a multiparameter identification method was proposed based on the PSO algorithm, which could achieve synchronous identification of multiple cable system parameters, including cable tension using measured multiorder frequencies.

### 5.1 Introduction to the PSO algorithm

The PSO algorithm, first introduced by Kennedy and Eberhart in 1995, is a swarm intelligent optimization algorithm (Kennedy and Eberhart 1995). This algorithm simulates the self-cognition and swarm communications of the bird flock during the predation process to realize the optimization of the minimum value of the objective function. The flight status of each particle in the swarm is described by a group of position vector and velocity vector, which, respectively, represent the possible solution to the problem and the update velocity of the solution. The particle, by referring to personal best positions and global best positions, constantly adjusts its flight velocity and position, thus searching for the optimal solution.

In the standard PSO algorithm, the iterative processes of the velocity and position of each particle in the swarm $X$ are given as follows:

$$v_{kj}(t+1) = \lambda_0 v_{kj}(t) + \lambda_1 \xi_1 \left( pbest_{kj}(t) - x_{kj}(t) \right) + \lambda_2 \xi_2 \left( gbest_j(t) - x_{kj}(t) \right) \tag{42}$$

$$x_{kj}(t+1) = x_{kj}(t) + v_{kj}(t+1) \tag{43}$$

where $v_{kj}(t)$ and $x_{kj}(t)$ are the velocity component and position component of the $j$th dimension at the $t$th generation for the $k$th particle, respectively; $\lambda_0$ is the inertia weight; $\lambda_1$ and $\lambda_2$ are the cognitive coefficient and social coefficient, respectively; $\xi_1$ and $\xi_2$ are the random numbers uniformly distributed on the interval [0, 1]; $pbest_{kj}(t)$ is the best position of the particle itself, that is, the personal best position passed by the $k$th particle in the $j$th dimension at the $t$th iteration; and $gbest_j(t)$ is the best position of the entire swarm, that is, the global best position passed by the swarm $X$ in the $j$th dimension at the $t$th iteration. The superiority or inferiority of the position of each particle is assessed by a fitness function. The latter is usually determined according to the optimization objective. The smaller the fitness value, the more superior the position of the particle and the better the solution it represents. The first term on the right side of Eq. (42) represents the inertia part, that is, which represents the influence of the particle's inertia on the current motion ; the second term is the cognition part, which represents the cognition and affirmation of the particle itself about its historical experience; and the third term is the social part, which represents the collaboration and knowledge sharing between the particles. The PSO algorithm uses these three parts to simulate the cooperative coevolution of the particle swarm, thus achieving the optimization.

The main parameters controlling the iterative update in the PSO algorithm are the inertia weight $\lambda_0$, the cognitive coefficient $\lambda_1$, and the social coefficient $\lambda_2$. The performance of PSO algorithm to the specific target may vary with different parameter combinations (Dan *et al.* 2018b). A reasonable choice of parameters can significantly improve the applicability and robustness of the PSO algorithm. The coefficients $\lambda_1$ and $\lambda_2$ control the influence of particle self-learning and swarm communications on the iterative update, respectively. The inertia weight $\lambda_0$ was first proposed by Shi and Eberhart (1998) to improve the convergence performance of the algorithm. This coefficient plays the role of balancing the global search and local search. Furthermore, in an attempt to improve the convergence performance of the algorithm, proposed the velocity update method using the constriction coefficient $\chi$ was proposed (Clerc 1999, Clerc and Kennedy 2002). Eq. (42) is rewritten in the following form:

$$v_{kj}(t+1) = \chi \left[ v_{kj}(t) + \varphi_1 \xi_1 \left( pbest_{kj}(t) - x_{kj}(t) \right) + \varphi_2 \xi_2 \left( gbest_j(t) - x_{kj}(t) \right) \right] \tag{44}$$

where

$$\chi = \frac{2}{\left| 2 - \varphi - \sqrt{\varphi(\varphi-4)} \right|}; \quad \varphi = \varphi_1 + \varphi_2; \quad \varphi > 4 \tag{45a, b, c}$$

Eberhart and Shi (2000) found that in most of the studies using the Clerc's approach, the parameters were usually configured as follows: $\varphi = 4.1$ and $\varphi_1 = \varphi_2$. They also compared the superiority or inferiority of the PSO algorithm in the form of Eq. (42) and Eq. (44). The PSO algorithm with the constriction coefficient was found to be superior. In the present study, the parameters were configured as $\varphi = 4.1$ and $\varphi_1 = \varphi_2$. Then, $\chi = 0.729$ and $\varphi_1 = \varphi_2 = 2.05$. If represented by Eq. (42), $\lambda_0$, $\lambda_1$, and $\lambda_2$ were, respectively, given as follows:

$$\lambda_0 = \chi = 0.729; \quad \lambda_1 = \lambda_2 = \frac{\varphi}{2} \times \chi = 1.494 \tag{46a, b}$$

## 5.2 Modified PSO algorithm for multiparameter identification

Multiple parameters of the cable system needed to be identified, and the approximate range of the cable parameters was large. In the present study, the boundary limit and velocity limit of particle flight in the conventional PSO algorithm were modified to ensure sufficient convergence and robustness of the PSO algorithm.

With the conventional PSO algorithm, the boundary limit for particles was imposed only during the initialization of the swarm *X*. That is, let the initial swarm be randomly, uniformly distributed within the initialization range, without setting up the flight boundary for each iteration separately. In engineering practice, the search range (solution space) of the cable parameters can be empirically determined, and the particles can be limited to fly in the solution space, thereby avoiding impractical cable parameters. A previous study (Robinson and Rahmat-Samii 2004) described three types of boundary conditions to limit the particle movement: absorbing walls, reflecting walls, and invisible walls. In this study, the absorbing walls were administered to ensure the stability of the algorithm without reducing the size of the particle swarm, obeying the following rules:

$$x_{kj} = Z_j; \quad v_{kj} = 0 \tag{47a, b}$$

where $Z_j$ is the boundary of the search space in the *j*th dimension. After the update of the *k*th particle, if its position component in the *j*th dimension exceeds the boundary $Z_j$, the position in the *j*th dimension of this particle is set on the boundary $Z_j$ and the particle energy is absorbed, that is, the velocity in the *j*th dimension is zeroed. Therefore, the particle is eventually pulled back toward the allowed solution space. Regarding the velocity limit of particle flight, the constriction coefficient $\chi$ played a certain role in the velocity limit of the particle flight according to previous studies (Clerc and Kennedy 2002). However, to further ensure the convergence performance of the algorithm, the maximum velocity $v_{kj,\max}$ of the *k*th particle in the *j*th dimension was configured to accelerate the convergence, and the value was half the length of the corresponding search interval.

The parameters of the cable system identified using the PSO algorithm in the present study were cable tension *H*, flexural stiffness *EI*, axial stiffness *EA*, rotational constraint stiffness parameters $K_{r1}$ and $K_{r2}$, and lateral support stiffness parameters $K_{s1}$ and $K_{s2}$ at the two ends of the cable. These seven parameters constituted the position vector of the particle $\boldsymbol{x}_k = \{H, EI, EA, K_{r1}, K_{r2}, K_{s1}, K_{s2}\}_k$. Introducing the position vector into the dynamic model of the cable system, the multiorder frequency $\boldsymbol{f}_{ck} = \{f_{c1}, f_{c2}, ..., f_{ci}, ..., f_{cn}\}_k$ was obtained, where $f_{ci} = \omega_i/(2\pi)$ is the frequency of the *i*th order. The difference between the calculated frequency $\boldsymbol{f}_c$ and the measured frequency $\boldsymbol{f}_m$ of the cable was narrowed progressively to obtain the parameters of the refined cable model closer to the real situation. To this end, the fitness function of the PSO algorithm was defined as the sum of squares of the difference between the calculated and measured frequencies of each order, that is,

$$F = \sum_{i=1}^{N}\left(f_{ci} - f_{mi}\right)^2 \tag{48}$$

where $f_{ci}$ is the calculated frequency of the *i*th order; $f_{mi}$ is the measured frequency of the *i*th order; and *N* is the number of measured frequencies. In engineering practice, the number of measured frequencies is usually less than seven. Thus, the number of measured frequencies is smaller than the number of parameters to be identified. In this case, the FBSU algorithm (Kim and Park 2007, Ma 2017) based on the Newton-Raphson method cannot be applied. However, the number of measured frequencies in the PSO algorithm has no rigorous limit, and discontinuity is also allowed between the orders of frequency. Therefore, the parameters can be identified with a small number of measured frequencies.

The workflow of using the PSO algorithm for multiparameter identification of the cable system is illustrated in Fig. 4, with the following steps:

(1) The known parameters of the cable system were input, including the cable length *L*, gravitational acceleration *g*, mass per unit length *m*, inclination angle *θ*, and number of cable segments *n* + 1 after discretization. The measured multiorder frequency $\boldsymbol{f}_m$ of the cable was input.

(2) The swarm was initialized. The number of dimensions of a particle *nvars* = 7; the population *size* = 100; the initial position of the particle was randomly and uniformly distributed within the initialization range; the initial velocity of a particle in each dimension was taken as 1/10 of the length of the corresponding search interval; the maximum number of iterations $t_{\max}$ and optimality tolerance *δ* were configured as appropriate. The initialization range of the particles could be the search range of each parameter of the cable, and the latter is usually determined empirically in engineering practice.

(3) The parameter vector represented by each particle was introduced into the dynamic model of the cable established in the present study. The value of fitness function $F_k$ for each particle was calculated. The personal best position $pbest_k$ of each particle and the global best position *gbest* in the current generation were determined according to the value of fitness function $F_k$ of each particle.

(4) The particle velocity was updated using Eq. (42), and a limit was imposed to prevent the particle velocity in any dimension from exceeding the corresponding maximum velocity $v_{kj,\max}$. The particle position was updated using Eq. (43), and the particles were prevented from exceeding the search boundary through the absorbing walls.

(5) Steps (3) and (4) were repeated until the fitness value $F_{gbest}$ of the global best position reached optimality tolerance $\delta$, or the number of iterations was $t_{\max}$. Then, the iteration was stopped.

(6) The global best position $gbest$ and best fitness value $F_{gbest}$ were obtained as output. and the group of cable parameters $\{H, EI, EA, K_{r1}, K_{r2}, K_{s1}, K_{s2}\}$ corresponding to $gbest$ were the results of multiparameter identification for the cable system.

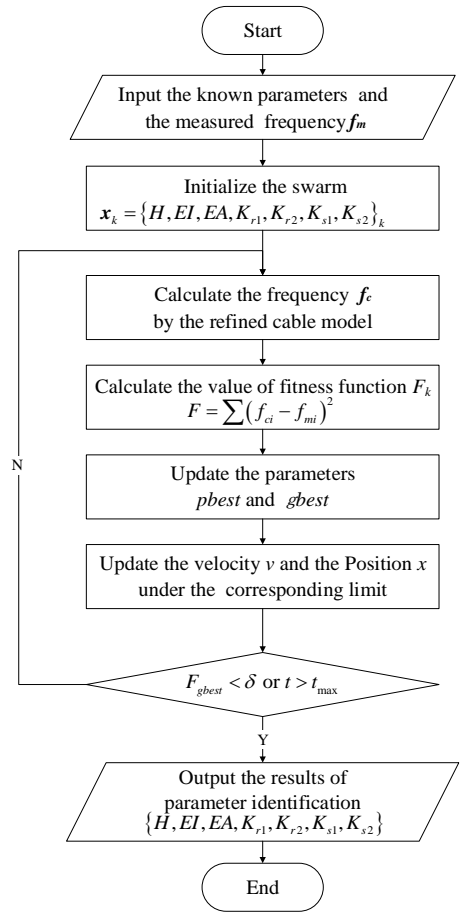

Fig. 4. Workflow of cable tension identification using the PSO algorithm.

## 6. Numerical case study

Four numerical cables with different properties were used to verify the feasibility of the proposed multiparameter identification method, as was commonly done in the existing studies (Kim and Park 2007, Ma 2017, Mebrabi and Tabatabai 1998). First, the physical parameters of the four numerical cables were configured and substituted into the refined cable model established in the present study. The multiorder frequency $f_m$ was calculated. Next, the calculated frequencies were taken as the exact value of frequencies. The parameters of each numerical cable were identified using the multiparameter synchronous identification method based on the PSO algorithm. Finally, the identified and configured values of the cable parameters were compared to verify the feasibility of the cable multiparameter identification method.

## 6.1 Numerical cable profiles

The mass of the four numerical cables per unit length was $m$ = 400 kg/m, the gravitational acceleration was $g$ = 9.8 N/kg, and the cable length was $L$ = 100 m. Other parameters are shown in Table 1. Furthermore, the cable inclination angle was uniformly set to 30º; and the boundary stiffness matrices $K_{r1}$, $K_{r2}$, $K_{s1}$, and $K_{s2}$ were uniformly set to $K_{r1} = K_{r2} = 2 \times 10^5$ N·m, $K_{s1} = 5 \times 10^5$ N/m, and $K_{s2} = 1 \times 10^6$ N/m.

In Table 1, the nondimensional parameter $\lambda^2$ is the Irvine parameter characterizing cable sag-extensibility and elastic features (Irvine 1981). The nondimensional parameter $\xi$ characterizes the flexural stiffness of the cable (Zui *et al.* 1996). These two parameters are, respectively, defined as follows:

$$\lambda^2 = \frac{LEA}{HL_e}\left(\frac{mgL}{H}\right)^2; \quad \xi = L\sqrt{\frac{H}{EI}} \tag{49a, b}$$

where

$$L_e = \int_0^L (ds/dx)^3 dx \approx L\left[1 + \frac{1}{8}\left(\frac{mgL}{H}\right)^2\right] \tag{50}$$

Thus, the larger the $\lambda^2$, the larger the sag-extensibility; the smaller the $\xi$, the larger the influence of flexural stiffness on cable vibration. As shown in Table 1, No. 1 represents a cable with small sag-extensibility and small flexural stiffness; No. 2 represents a cable with large sag-extensibility and small flexural stiffness; No. 3 represents a cable with small sag-extensibility and large flexural stiffness; and No. 4 represents a cable with large sag-extensibility and large flexural stiffness. The tension, flexural stiffness, and sag-extensibility of the four cables differed from one another, and so they could cover a diversity of cable features in engineering practice.

Table 1. Physical parameters of each numerical cable.

| Cable | $\lambda^2$ | $\xi$ | $H$ ($10^6$N) | $E$ (Pa) | $A$ (m$^2$) | $I$ (m$^4$) |
|---|---|---|---|---|---|---|
| No.1 | 0.79 | 605.5 | 2.9036 | 1.5988E+10 | 7.8507E-03 | 4.9535E-06 |
| No.2 | 50.70 | 302.7 | 0.7259 | 1.7186E+10 | 7.611 E-03 | 4.6097E-06 |
| No.3 | 1.41 | 50.5 | 26.13254 | 2.0826E+13 | 7.8633E-03 | 4.9204E-06 |
| No.4 | 50.70 | 50.5 | 0.7259 | 4.7834E+8 | 0.7345E-01 | 5.9506E-03 |

In the finite difference model of the cable, the cable was divided into 100 segments along the chordwise direction ($n$ = 99 and $a$ = 1 m). The whole process of cable tension identification was carried out in the MATLAB platform. The physical parameters of the numerical cables were substituted into the dynamic cable model. In previous studies (Kim and Park 2007, Ma 2017), at least an equal number of measured frequencies were needed to identify all parameters synchronously. Seven parameters were needed to be identified in the present study, and therefore the frequencies of the first seven orders were calculated. The calculated frequencies of the first seven orders for each cable are shown in Table 2.

Table 2. Exact values of frequencies of the first seven orders for the numerical cables (Hz).

| Cable | 1st | 2nd | 3rd | 4th | 5th | 6th | 7th |
|---|---|---|---|---|---|---|---|
| No. 1 | 0.4229 | 0.8267 | 1.2404 | 1.6541 | 2.0681 | 2.4824 | 2.8970 |
| No. 2 | 0.4120 | 0.4308 | 0.6487 | 0.8498 | 1.0626 | 1.2749 | 1.4876 |
| No. 3 | 0.8793 | 1.7964 | 2.9918 | 4.2928 | 5.6725 | 7.1265 | 8.6589 |
| No. 4 | 0.4151 | 0.4306 | 0.6538 | 0.8636 | 1.0977 | 1.3411 | 1.5981 |

As shown in Table 2, the first-order and second-order frequencies are very close for No. 2 and 4 cables with large sag-extensibility. This result convincingly suggested that sag-extensibility has a major impact on cable frequencies, especially lower-order symmetric frequencies. According to the calculation results of No. 1 and 2 numerical cables, the influence of flexural stiffness on frequency is mainly manifested in higher-order frequencies. For No. 3 cable with a large flexural stiffness, the differences between higher-order frequencies are significantly larger than those between lower-order frequencies. However, the frequency difference of No. 1 cable with small flexural stiffness has no significant change. In addition, as No. 3 cable has a smaller lateral support stiffness relative to its large flexural stiffness, the cable frequencies are generally lower than those with fixed boundary condition.

## 6.2 Numerical validation of multiparameter identification

The calculated frequencies presented in Table 2 were used as the exact frequency values $f_m$ for each numerical cable. Parameters of each numerical cable were identified using the multiparameter synchronous identification method based on the PSO algorithm. For numerical cables with theoretical exact values $x^*$ of parameters $x$, the search ranges of parameters $H$, $EI$, and $EA$ were set to $[0.75H^*, 1.25H^*]$, $[0.75EI^*, 1.25EI^*]$, and $[0.75EA^*, 1.25EA^*]$, respectively. The search ranges of boundary stiffness parameters were set to $[0.5K_r^*, 1.5K_r^*]$ and $[0.5K_s^*, 1.5K_s^*]$, respectively. The maximum number of iterations was set to $t_{max} = 200$, without configuring optimality tolerance $\delta$.

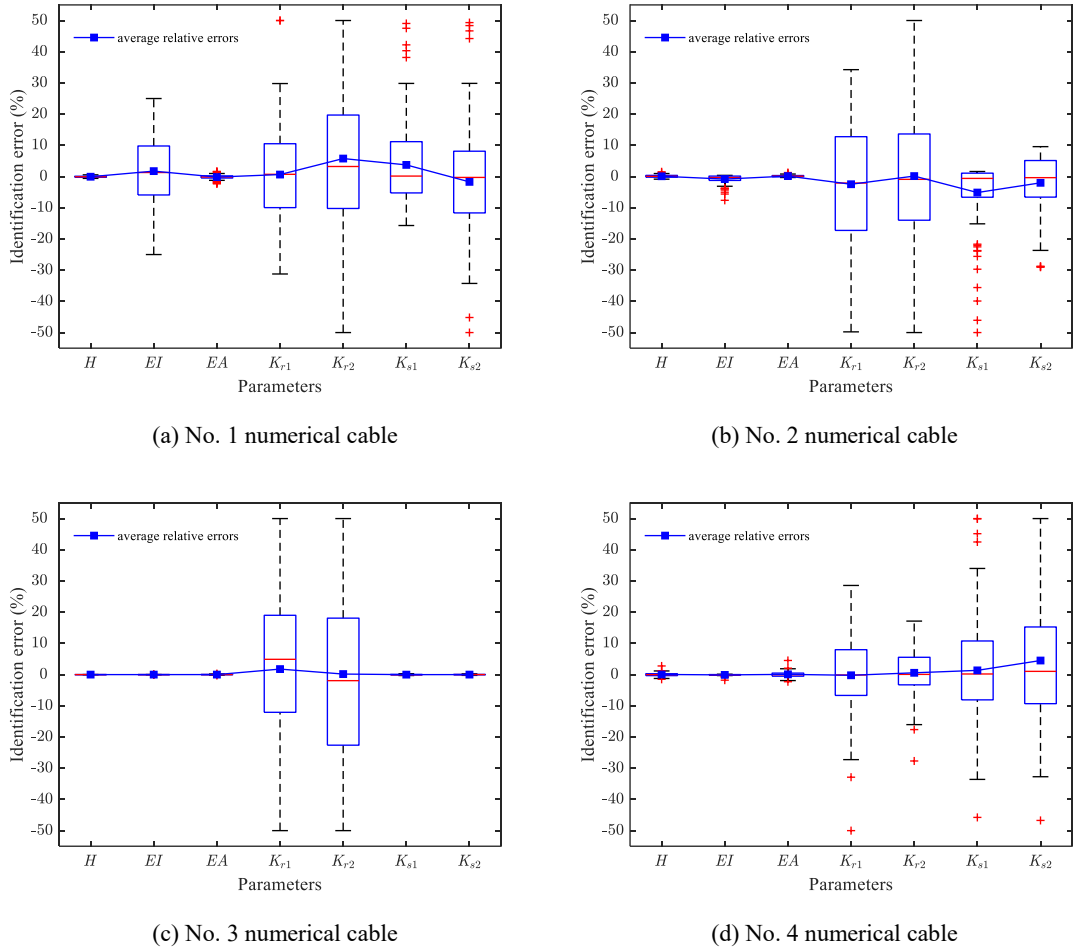

(a) No. 1 numerical cable  (b) No. 2 numerical cable

(c) No. 3 numerical cable  (d) No. 4 numerical cable

Fig. 5. Relative error boxplots of 100 independent seven-parameter identifications for numerical cables.

The identification results of each calculation were not entirely consistent because of the stochastic nature of the PSO algorithm. So the calculation was run 100 times independently for each numerical cable, and the final parameter identification results were determined based on statistical principles. The deviations between the parameter identification results of 100 calculations and the exact values were represented in boxplots, as shown in Fig. 5. The *y*-axis represents the relative errors of the parameter identification results, while the *x*-axis represents each parameter of the cable system. The rectangular points represent the mean relative errors of parameter identification results.

Generally speaking, the degree of dispersion is relatively small for the identification results of *H*, *EI*, and *EA*, and the identified values of each calculation are basically consistent, while the difference of the identified values for the boundary stiffness of each calculation is slightly large. However, the identified values of boundary stiffness are not randomly and uniformly distributed within the search range, but in a concentrated manner near the mean values. Based on this result, the mean identified values of each calculation were taken as the final identified values of the parameters and are listed in Table 3.

Table 3. Parameter identification results of numerical cables.

| Cable | | $H$ (kN) | $EI$ ($10^4$N·m$^2$) | $EA$ ($10^8$N) | $K_{r1}$ ($10^5$N·m) | $K_{r2}$ ($10^5$N·m) | $K_{s1}$ ($10^5$N/m) | $K_{s2}$ ($10^5$N/m) |
|---|---|---|---|---|---|---|---|---|
| | Exact values | 2903.6 | 7.9197 | 1.2552 | 2.00 | 2.00 | 5.00 | 10.00 |
| No. 1 | Estimation values | 2901.9 | 8.0571 | 1.2530 | 2.01 | 2.12 | 5.19 | 9.83 |
| | Error (%) | −0.06 | −1.74 | −0.17 | 0.64 | 5.76 | 3.72 | −1.71 |
| | Exact values | 725.9 | 79222 | 1.3080 | 2.00 | 2.00 | 5.00 | 10.00 |
| No. 2 | Estimation values | 726.6 | 78579 | 1.3100 | 1.95 | 2.00 | 4.74 | 9.80 |
| | Error (%) | 0.09 | −0.81 | 0.15 | −2.47 | 0.15 | −5.13 | −2.01 |
| | Exact values | 26132.54 | 10247 | 1637.6 | 2.00 | 2.00 | 5.00 | 10.00 |
| No. 3 | Estimation values | 26132.53 | 10247 | 1637.7 | 2.04 | 2.00 | 5.00 | 10.00 |
| | Error (%) | −0.00003 | 0.0001 | 0.006 | 1.77 | 0.17 | −0.01 | 0.001 |
| | Exact values | 725.9 | 284.64 | 1.3080 | 2.00 | 2.00 | 5.00 | 10.00 |
| No. 4 | Estimation values | 726.2 | 284.37 | 1.3090 | 2.00 | 2.01 | 5.07 | 10.45 |
| | Error (%) | 0.05 | −0.09 | 0.08 | −0.22 | 0.57 | 1.36 | 4.51 |

Table 3 shows that the tension identification of numerical cables could reach a precision of 0.1% by considering the means of 100 calculations as the final results. As to the identification of *EI* and *EA*, the relative identification error is controlled to less than 2%. Even for the identified boundary stiffness with a higher degree of dispersion, the relative identification error is controlled to less than 6%. The results show that the proposed multiparameter identification method has a good identification effect for all four numerical cables.

Table 4. Results of cable tension identification using the five methods (kN).

| Cable | No.1 | No.2 | No.3 | No.4 |
|---|---|---|---|---|
| Exact tension | 2,903.6 | 725.9 | 26,132.54 | 725.9 |
| **Proposed approach** | **2,901.9 (–0.06)** | **726.6 (0.09)** | **26,132.53 (–0.00003)** | **726.2 (0.05)** |
| String theory in Eq. (51) | 2,754.8 (–5.12) | 1,013.7 (39.65) | 18,180.87 (–30.43) | 1,058.2 (45.78) |
| Axially loaded beam theory in Eq. (52) | 2,778.0 (–4.33) | 1,423.8 (96.15) | 13,023.34 (–50.16) | 1,428.6 (96.81) |
| The empirical formula (Zui *et al*. 1996) | 2,713.5 (–6.55) | 731.7 (0.80) | 10,977.31 (–57.99) | 677.3 (–6.67) |
| FBSU algorithm | Divergence | Divergence | Divergence | Divergence |

Note: (·) is % error.

Table 4 presents the identified cable tensions of four numerical cables using four different identification methods compared with those using the proposed method. (1) The string theory could calculate the cable tension from the frequency of any order. The final identified tension was determined by the mean of cable tensions calculated from the frequency of each order, as represented by Eq. (51). (2) Based on the axially loaded beam theory, the frequency-based method considering flexural stiffness correction was written as Eq. (52), and linear regression was performed on this equation using multiorder frequencies to identify cable tensions. (3) The empirical formula proposed by Zui *et al*. (1996) used the second-order frequency for cable tension identification. (4) Combined with the refined cable model and considering the complex boundary stiffness proposed in the present study, the FBSU algorithm based on the Newton-Raphson method in previous studies (Kim and Park 2007, Ma 2017) was modified. Then, this method was applied to numerical cases of multiparameter identification. The initial values of the parameters used in FBSU algorithm were [$1.25H^*$, $1.25EI^*$, $1.25EA^*$, $1.5K_{r1}^*$, $1.5K_{r2}^*$, $1.5K_{s1}^*$, $1.5K_{s2}^*$].

$$H = \frac{\sum_{i=1}^{N} 4mL^2 \left(\frac{f_i}{i}\right)^2}{N} \tag{51}$$

$$\left(\frac{f_i}{i}\right)^2 = \left(\frac{EI\pi^2}{4mL^4}\right)i^2 + \frac{H}{4mL^2} \tag{52}$$

The comparison shows that the approach of cable tension identification in this study is much more accurate than the other methods. The errors of the identified tensions in the four numerical cables is all controlled to less than 0.1%. Thus, the present method is considered applicable to most cables in engineering practice. Except for the relative error of 5% in No. 1 cable using the string theory, large errors are observed in all other cables using this method. The reason is that the string theory did not fully consider the influence of flexural stiffness, axial stiffness, and complex boundary conditions on cable vibration. The axially loaded beam theory only produces an improved identification of cable tension relative to the string theory in No. 1 cable. However, for the remaining three cables, the identification error rather increases because the identification errors in the other three cables are primarily attributed to the sag-extensibility and complex boundary conditions. Therefore, considering the flexural stiffness alone could not achieve an effective correction of identification results. The tension identification errors for Nos. 1, 2, and 4 cables using the empirical formula proposed by Zui *et al.* (1996) are controlled to less than 7%, but the error is large for No. 3 cable. Therefore, this empirical formula can eliminate the influence of sag-extensibility and flexural stiffness on cable tension identification, but can not effectively reduce the error of cable tension identification caused by boundary stiffness. The factors considered in cable tension identification using the four methods are shown in Table 5.

Table 5. Influencing factors for cable tension considered in each of the four methods.

| Influencing factor | Flexural stiffness | Sag-extensibility | Boundary stiffness |
|---|---|---|---|
| **Proposed approach** | ✓ | ✓ | ✓ |
| String theory in Eq. (51) | – | – | – |
| Axially loaded beam theory in Eq. (52) | ✓ | – | – |
| The empirical formula (Zui *et al*. 1996) | ✓ | ✓ | – |

Also, the FBSU algorithm based on the Newton-Raphson method has a heavy reliance on the choice of initial values. This algorithm could not converge under the given initial values and failed to yield valid parameter identification results. In contrast, the proposed method has a less strict requirement on the choice of initial values. Moreover, by fully considering the influence of complex boundary stiffness, sag-extensibility, and flexural stiffness, this method has a widened scope of application, improves the accuracy of cable tension identification, and produces a better parameter identification result.

### 6.3 Influence of frequency quantity on identification accuracy

As analyzed earlier, the cable tension identification method proposed in the present study achieved a good cable tension identification with the frequencies of the first seven orders known. However, in engineering practice, it is usually hard to identify the frequencies of all seven orders used for cable parameter identification. To verify the applicability of the proposed method using a fewer number of known frequencies, which is usually the case in real-world scenarios, the number of frequencies gradually decreased from higher orders to lower orders, based on the initial seven orders of frequencies and changes in cable tension identification results using the proposed approach were detected during this process. Except for the varying number of frequencies during the calculation, the other parameters were kept unchanged. For each condition, the calculation was run 100 times, and the means of cable tension were used as the final identified values of the cable tension. The calculation results are given in Fig. 6.

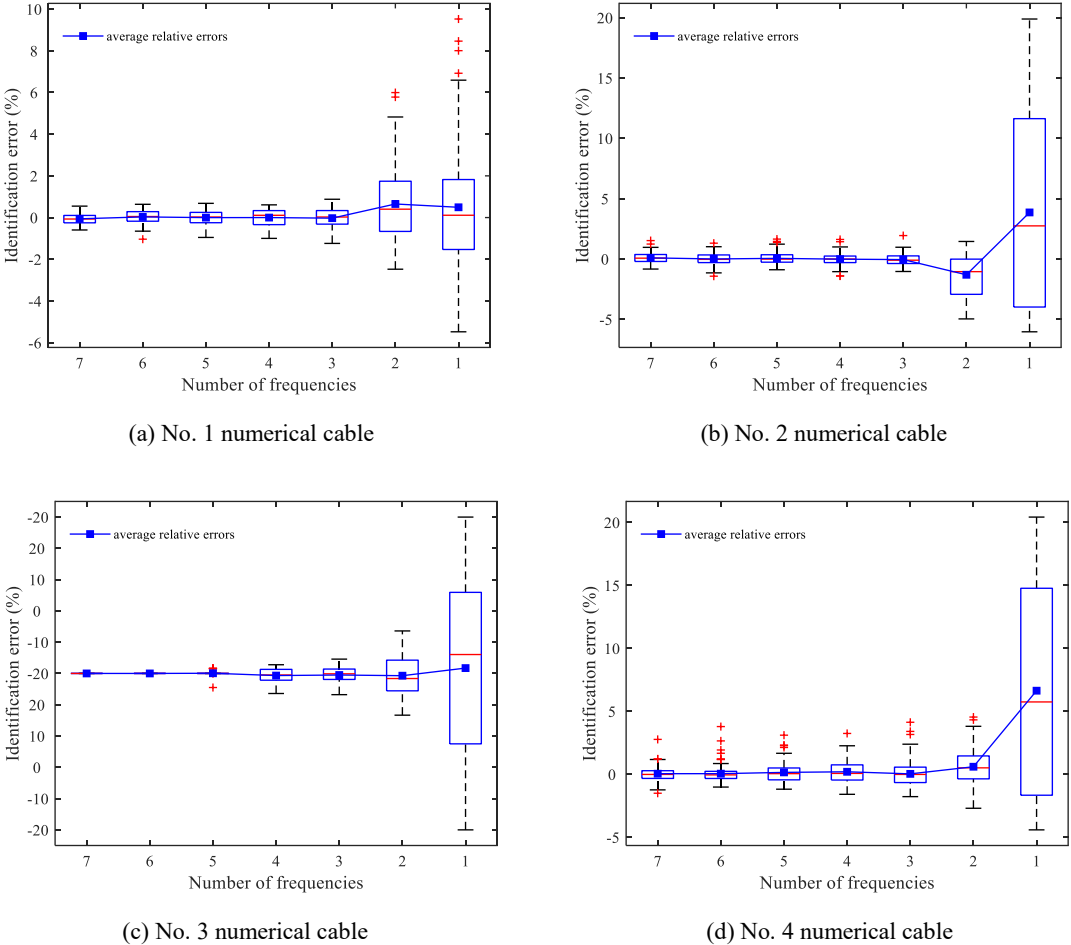

(a) No. 1 numerical cable  (b) No. 2 numerical cable

(c) No. 3 numerical cable  (d) No. 4 numerical cable

Fig. 6. Relative error boxplots of 100 cable tension identifications with different numbers of known frequencies for numerical cables

As shown in Fig. 6, the degree of dispersion of errors in cable tension identification increases as the number of frequencies decreases for the four numerical cables. When the number of frequencies is more than 3, the degree of dispersion of cable tension identification results changes but not significantly, moreover, a precision of less than 1% is always maintained for identified values of cable tension. When the number of frequencies is 2, the degree of dispersion of cable tension identification results increases suddenly, and a more significant deviation is found in the means of identification results in each numerical cable. When only the fundamental frequency is present, the degree of dispersion of cable tension identification results further increases. The maximum deviation of a single identification result of cable tension exceeds by 20%. Especially for No. 4 cable, the mean error of identification also exceeds by 6%, indicating the failure of the identification method using the fundamental frequency only because the influence of the flexural stiffness on frequency can not be reflected well in the absence of high-order frequencies. Also, given the influence of sag-extensibility on symmetric modal frequencies of lower orders, it is difficult to identify cable tensions based on the fundamental frequency directly.

In summary, the first three orders frequencies of cable vibration contain the majority of information about cable vibration. Thus, the proposed method can achieve accurate cable tension identification with at least the frequencies of the first three orders known, considering six parameters, including flexural stiffness, axial stiffness, as well as rotational constraint stiffness and lateral support stiffness at the two ends of the cable.

## 7. Engineering case study

### 7.1 Background

Jindong Bridge is located in Dongchuan District of Kunming City, Yunnan Province, China, and it is a suspension bridge with the largest span over the Jinsha River. As shown in Fig. 7, the bridge has a steel truss girder with a single main span of 730 m, bridge deck width of 20 m, and center distance between the upstream and downstream main cables of 17.5 m. The main cables of the two side spans are asymmetrically arranged. The main cables are made of prefabricated parallel steel wire strands. Each tower has a portal frame structure.

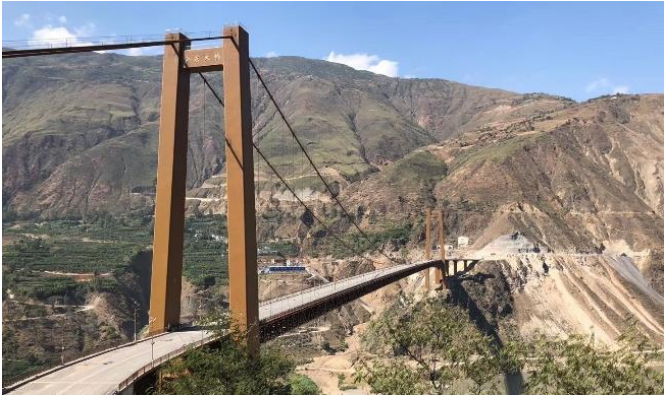

Fig. 7. Jindong Bridge.

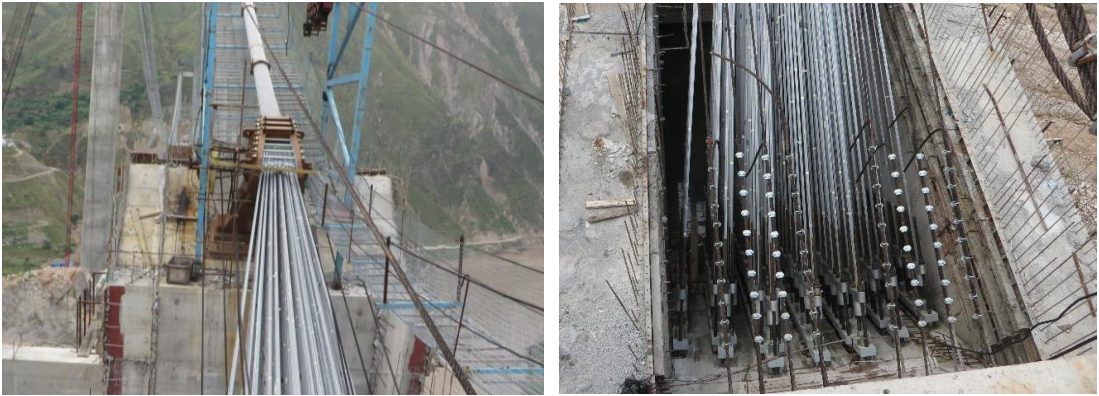

(a) Splay saddle of anchor span strands      (b) Front anchor facet of anchor span strands

Fig. 8. Anchor span strands of the main cable in Jindong Bridge.

Fig. 8 shows the anchor span strands of the main cable in Jindong Bridge. The anchor system is the prestressed anchor system, and the anchor span strands are anchored in the cantilever linkage of the front anchor facet. The vibration of the anchor span strand is influenced by not only sag-extensibility and flexural stiffness but also complex boundary condition constraints. The upper end of the anchor span strand is stacked with the other strands in the saddle groove, and the lower end is connected to the anchor head of the prestressed steel strand in the front anchor facet via the cantilever linkage. When the anchor span strands vibrate, elastic rotation and elastic lateral displacement occur on boundaries, instead of simply hinged or consolidated boundary conditions. Therefore, if the conventional frequency-based method built upon the string theory is used, that is, Eq. (51), a considerable deviation of the obtained cable tension from the true cable tension may occur during the tension control of the anchor span strands. In the process of the cable tension control, cable tensions measured by pressure rings are generally considered accurate. In this study, the proposed method was applied to the tension measurement of anchor span strands in Jindong Bridge, and the results were compared with those from the pressure rings.

The frequencies of the anchor span strands were measured using a dynamic tester. Fig. 9 shows the layout of the pressure rings in the anchor span strands. Table 6 shows the physical parameters of the anchor span strands ① and ② in Jindong Bridge. Table 7 shows the frequencies of the anchor span strands ① and ② measured in the tension control, the cable tension $H_f$ calculated by the conventional frequency-based method, and the cable tension $H_p$ obtained from the pressure ring.

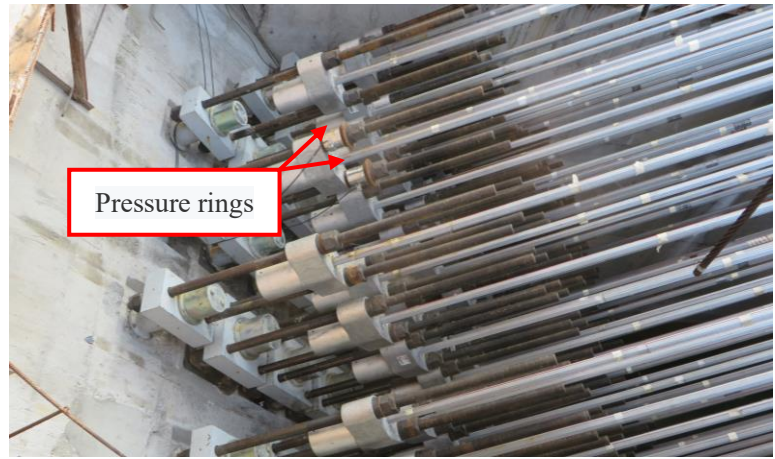

Fig. 9. Layout of pressure rings in the anchor span strand of the main cable.

Table 6. Physical parameters of anchor span strands.

| Cable | $m$ (kg/m) | $g$ (N/kg) | $L$ (m) | $\theta$ (°) | $E$ (Pa) | $A$ (m$^2$) | $I$ (m$^4$) |
|---|---|---|---|---|---|---|---|
| ① | 15.17077 | 9.8 | 18.884 | 31.35 | 1.97E+11 | 1.932582 E-03 | 3.3059E-07 |
| ② | 15.17077 | 9.8 | 17.936 | 39.88 | 1.97E+11 | 1.932582 E-03 | 3.3059E-07 |

Table 7. Measured frequencies and cable tensions of the anchor span strands.

| Cable | $f_{m1}$ (Hz) | $f_{m2}$ (Hz) | $f_{m3}$ (Hz) | $H_f$ (kN) in Eq. (51) | $H_p$ (kN) |
|---|---|---|---|---|---|
| ① | 2.990 | 5.882 | 8.896 | 190.31 | 174.19 |
| ② | 2.838 | 5.558 | 8.459 | 154.40 | 131.83 |

As shown in Table 7, field measurements obtained only the frequencies of the first three orders of the anchor span strands. A considerable difference was found between the cable tension calculated by the conventional frequency-based method, and that measured by the pressure ring. The findings indicated that the conventional frequency-based method for tension measurement of anchor span strands was not reliable.

## 7.2 Multiparameter identification via the PSO algorithm

Known parameters of anchor span strands were used in the proposed method of cable parameter identification. Synchronous identification of cable tension and other system parameters was achieved according to measured frequencies. In the finite difference model, the number of internal nodes was $n = 99$, which divided the strand into 100 segments. No optimality tolerance $\delta$ was set up in PSO algorithm, and the number of iterations $t_{max} = 200$.

Due to the fact that the approximate ranges of the cable parameters often can be pre-estimated (Xie and Li 2014), the method for determining search ranges of each parameter in this paper was as follows: The cable tension $H$ was generally close to the measured value $H_f$ using the frequency-based method. Thus, the search interval for the cable tension was $[0.5H_f, 1.5H_f]$. For $EI$ and $EA$, the parameters $EI_{cal}$ and $EA_{cal}$ could be obtained by using the axially loaded beam theory. The search intervals were set to $[0.2EI_{cal}, 2EI_{cal}]$ and $[0.2EA_{cal}, 2EA_{cal}]$, respectively. However, it was difficult to determine a precise search range for boundary stiffness due to the lack of prior knowledge. The search range of the boundary stiffness of anchor span strands was determined by parameter analysis of cable boundary stiffness, which were realized by substituting the physical parameters of cable ① in Table 6 and $H_f$ of cable ① in Table 7 into the dynamic model of the cable, assuming that $K_{r1} = K_{r2} = K_r$ and $K_{s1} = K_{s2} = K_s$. Taking the determination of the research range for $K_s$ as an example, Fig. 10 shows the relationship between $K_s$ and cable frequencies of the first three orders under different $K_r$.

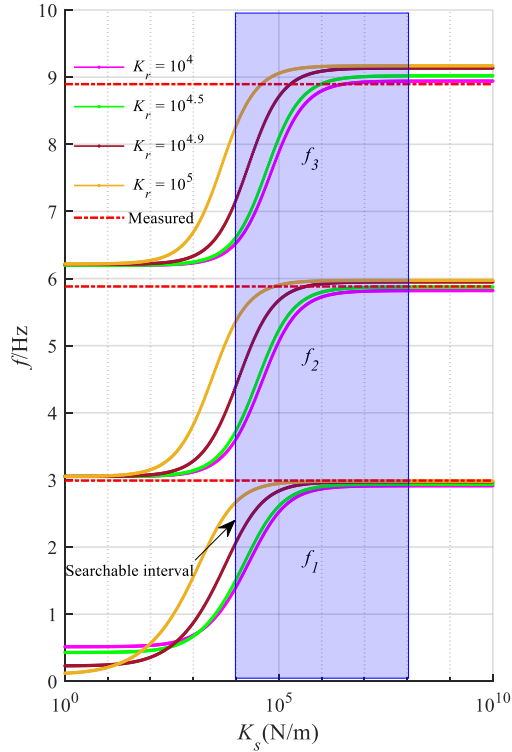

Fig. 10. Relationship between lateral support stiffness $K_s$ and frequency.

As shown in Fig. 10, $K_s$ corresponding to the measured frequency $f_m$ under different $K_r$ was generally within the higher interval $[10^4, 10^8]$. When $K_s < 10^4$, the measured frequency did not intersect with the calculated frequency; when $K_s > 10^8$, the frequencies basically remained unchanged, and the lateral support stiffness was considered infinitely large. Therefore, the search interval for $K_s$ was set to $[10^4, 10^8]$. Similarly, the search range for $K_r$ was set to $[10^4, 10^6]$. Then, the proposed multiparameter identification method was run 100 times.

The relative differences of the result by each calculation from the average in the cable ① and ② were presented in boxplots, as shown in Fig. 11. The cable tension $H$, flexural stiffness $EI$, and axial stiffness $EA$ identified by each independent calculation are generally consistent, and the identification results can be expressed by the means. However, the interquartile range of boxplot for boundary stiffness is significantly larger than that for $H$, $EI$, and $EA$, and the identified values are distributed over the entire search range. Therefore, the identification results should not be described by the means. The identified values of cable tension $H$, flexural stiffness $EI$, and axial stiffness $EA$, as well as the exact value of cable tension $H_p$ measured from the pressure ring and the calculated values of flexural stiffness $EI_{cal}$ and axial stiffness $EA_{cal}$ using the axially loaded beam theory, are listed in Table 8.

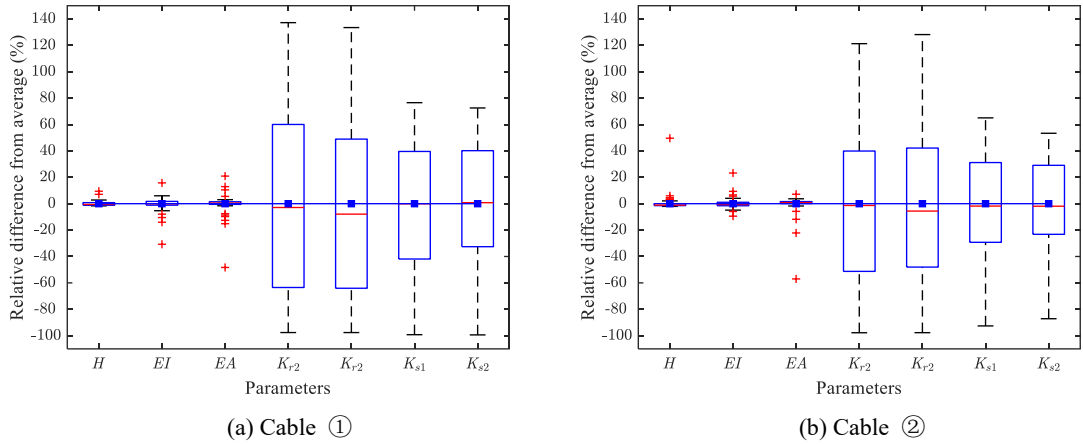

(a) Cable ①  (b) Cable ②

Fig. 11. Boxplots of 100 seven-parameter identifications for engineering cables

Table 8. Identified parameter values of the anchor span strands and the corresponding exact or calculated values.

| Cable | | $H$ (kN) | $EI$ (N·m$^2$) | $EA$ (N) |
|---|---|---|---|---|
| ① | Exact values | 174.19 | — | — |
| | Calculated values | — | 67,126.28145 | 380,718,680.9 |
| | Estimation results | 173.99 | 21,372.67584 | 576,262,658.9 |
| ② | Exact values | 131.83 | — | — |
| | Calculated values | — | 67,126.28145 | 380,718,680.9 |
| | Estimation results | 136.10 | 27,060.60398 | 561,853,249.9 |

As shown in Table 8, the identified cable tension $H$ is close to that identified by the pressure ring, indicating that the proposed method can improve the accuracy of cable tension identification relative to the conventional frequency-based method. The identified values of flexural stiffness $EI$ and axial stiffness $EA$ differs significantly from the values of $EI_{cal}$ and $EA_{cal}$ calculated by the axially loaded beam theory. The identified value of $EI$ is always smaller than the calculated value because the shearing and bending mechanisms of cables were different from those of the beams. The looseness and slippage of each wire in the cable reduces the flexural stiffness $EI$ of the cable. Therefore, the flexural stiffness calculated from the axially loaded beam theory is not the flexural stiffness in actual cable vibration. The identified value of $EA$ is larger than $EA_{cal}$ and might be the impractical value. However, the proposed approach has a good identification capacity on axial stiffness as shown by numerical validation. This phenomenon in engineering cases might be because the sag-extensibility of the anchor span strands is small and the nonlinear stiffness matrix $\boldsymbol{K}_1$ contributes little to the total stiffness matrix $\boldsymbol{K}$. Kim and Park (2007) obtained similar results using the FBSU algorithm based on the Newton-Raphson method for parameter identification and pointed out that the identified value of $EA$ had no impact on the accuracy of cable tension $H$ in this situation. Furthermore, some analysis on the axial vibration may be necessary to estimate the accurate value of *EA* in actual cable vibration.

Similarly, the cable tension identification results using the present approach in the cables ① and ② were compared with those of the other four methods to verify the superiority of the proposed cable tension identification method. These four methods were as follows: (1) Eq. (51) based on the string theory. (2) Eq. (52) based on the axially loaded beam theory. Given the small number of frequencies, the linear regression result of Eq. (52) was usually unacceptable. Therefore, flexural stiffness *EI* was substituted into Eq. (52) as a known parameter rather than derived from linear regression, and the value of *EI* was taken as the calculated value $EI_{cal}$. (3) The empirical formula (Zui *et al*. 1996) used the second-order frequency for cable tension identification. (4) The FBSU algorithm based on the Newton-Raphson method. The cable tensions identified by different methods are given in Table 9.

Table 9. Cable tension identification results using the five methods (kN).

| Cable | ① | ② |
|---|---|---|
| Measured tension ($H_p$) | 174.19 | 131.83 |
| **Proposed approach** | **173.99 (−0.01)** | **136.10 (3.24)** |
| String theory in Eq. (51) ($H_f$) | 190.31 (9.25) | 154.40 (17.12) |
| Axially loaded beam theory in Eq. (52) | 181.29 (4.08) | 147.36 (11.78) |
| The empirical formula (Zui *et al*. 1996) | 154.92 (−11.06) | 119.45 (−9.39) |
| FBSU algorithm | Divergence | Divergence |

Note: (·) is % error.

Table 9 shows that the difference between the cable tensions calculated by the string theory and the tensions measured by the pressure ring is more than 9%. Also, irrespective of cable ① or ②, the axially loaded beam theory shows some improvement in terms of cable tension identification compared with the string theory, indicating that taking the influence of the flexural stiffness into account does improve the accuracy of cable tension identification, but still fails to control the error of cable tension identification both in cables ① and ② to less than 10%. The cable tension obtained from the empirical formula (Zui *et al*. 1996) is different by about 10% compared with the measured value. Therefore, an accurate cable tension identification result can not be obtained by using the empirical formula (Zui *et al*. 1996), either. The FBSU algorithm can not identify cable parameters due to the insufficient number of measured frequencies. However, the proposed method reduces the error of cable tension identification in both cables to less than 4%, and the error in cable ② is slightly larger than that in cable ①. The comparison shows that the tension identification accuracy in the anchor span strands using the proposed method is significantly higher than that using the conventional frequency-based method, frequency-based method with flexural stiffness correction, and the empirical formula proposed by Zui *et al*. (1996), because the proposed method takes into account the influence of boundary stiffness on cable vibration as well as the cable flexural stiffness and sag-extensibility. Additionally, the proposed method has fewer requirements for measured frequencies compared with the FBSU algorithm, and hence the applicability was higher.

## 7.3 Discussion

### 7.3.1 Influence of *EA*'s identification result

During the parameter identification of cables ① and ②, the axial stiffness *EA* was overestimated considerably, which was contrary to the expectation of this study. Thus, the tension identification results of cables ① and ② using the calculated value $EA_{cal}$ as the known axial stiffness *EA* were compared with those considering *EA* as the unknown parameter, and the findings are listed in Table 10.

Table 10. Cable tension identification results using the present approach in two ways.

| Cable | ① | | ② | |
|---|---|---|---|---|
| | $H$ (kN) | $F$ | $H$ (kN) | $F$ |
| Exact tension | 174.19 | — | 131.83 | — |
| Unknown $EA$ | 173.99 (−0.01) | <10$^{-4}$ | 136.10 (3.24) | <10$^{-4}$ |
| Given $EA$ | 172.83 (−0.78) | <0.0003 | 138.13 (4.78) | <0.0005 |

Note: (·) is % error; $F$ represents the value of fitness function (error function).

Table 10 shows that the cable tension identified from the known parameter $EA$ is close to that identified by treating $EA$ as an unknown parameter. The major difference is the error $F$ between the calculated and measured frequencies, indicating that finding a group of cable parameters with a complete correspondence to measured frequencies is impossible with the given axial stiffness $EA_{cal}$. In addition, the identified cable tension has a larger error when the axial stiffness is taken as $EA_{cal}$. Therefore, it is considered that the identification of $EA$ has no negative impact on the tension identification accuracy in the cables ① and ②.

**7.3.2 Reasons for multiple solutions of boundary stiffness**

Furthermore, it is also difficult to identify the boundary stiffness in engineering cases. The possible reasons are as follows: (1) The number of the measured frequencies is small, resulting in nonuniqueness of boundary stiffness combinations for the corresponding frequencies; and (2) the search range of boundary stiffness parameters is too large. Therefore, a parameter analysis is necessary for the boundary stiffness. A cable system consists of four unknown boundary stiffness parameters, including the rotational constraint stiffness and lateral support stiffness at the two ends of the cable. Some further assumptions are needed to complete the parameter analysis due to a large number of parameters. Based on the cable boundary conditions in engineering practice, the following two scenarios are assumed and the analysis is conducted separately:

(1) Scenario 1: Boundary conditions at the two ends of the cable are consistent, that is, $K_{r1} = K_{r2}$ and $K_{s1} = K_{s2}$.

Taking cable ① as the analysis case, the geometric and physical parameters in Table 6 and the identified values of $H$, $EI$, and $EA$ were substituted into the free vibration model of the cable in this study. The values of the boundary stiffnesses $K_r$ and $K_s$ at the two ends of the cable lay within the interval [0, 10$^{10}$]. The curve of frequency with boundary stiffness is shown in Fig. 12(a), where the coordinate axis scale of boundary stiffness is the logarithmic scale.

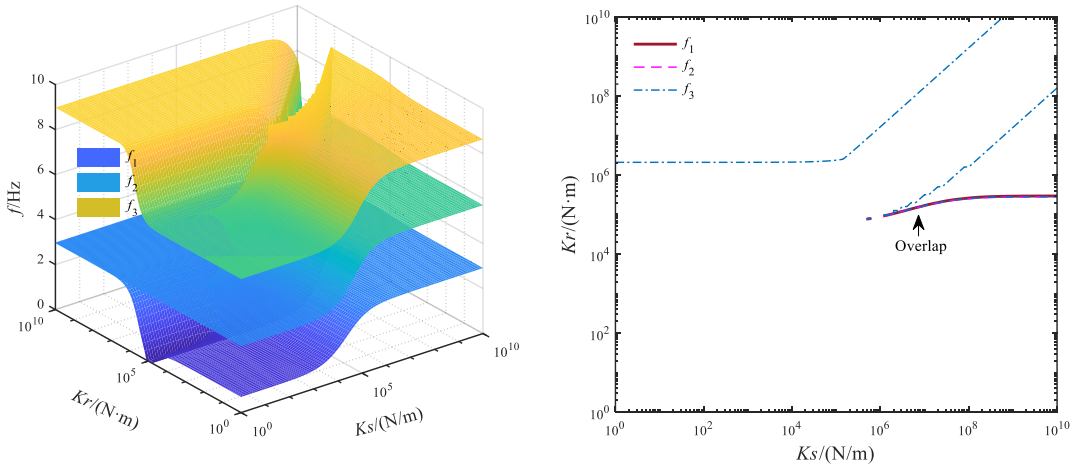

(a) Curve of multiorder frequencies of cable ① with boundary stiffness    (b) Isolines of multiorder frequencies of cable ①

Fig. 12. Parameter analysis for boundary of stiffness scenario 1 ($K_{r1} = K_{r2}$, $K_{s1} = K_{s2}$).

As shown in Fig. 12(a), the frequency of each order increases with $K_s$. Moreover, $K_s$ has a considerable impact on the changes of frequencies within the interval [$10^3$, $10^6$]. The influence of $K_r$ on frequencies increases with the increase in $K_s$, indicating that for the cable with weaker lateral support, rotational constraint stiffness is not the primary condition controlling frequencies and its influence on frequencies can be negligible. The previous studies discussing the influence of rotational stiffness on frequency were generally conducted under the condition where the lateral support was infinitely large. Therefore, whether considering rotational stiffness or not leads to a large difference in the result. However, when the lateral support stiffness is uncertain, the influence of rotational constraint stiffness and lateral support stiffness on the cable frequency should be considered comprehensively. Based on the frequencies of the first three orders measured for cable ①, the isolines for boundary stiffness are shown in Fig. 12(b). The boundary stiffnesses $K_r$ and $K_s$ corresponding to frequencies of each order overlap with each other, rather than intersecting at one point. Thus, under the aforementioned scenario, the boundary stiffness has multiple solutions. The unique appropriate values of boundary stiffness cannot be obtained using the frequencies of the first three orders alone.

(2) Scenario 2: The lateral support stiffness is infinitely large at one end, and the rotational constraint stiffness is infinitely large at the other end, that is, $K_{s1} = \infty$ and $K_{r2} = \infty$.

Similar to scenario 1, the geometric and physical parameters and the identified values of $H$, $EI$, and $EA$ of cable ① are substituted into the free vibration model of the cable in this study. Let $K_{s1} = \infty$ and $K_{r2} = \infty$. The values of the other boundary stiffness $K_{r1}$ and $K_{s2}$ of the cable lay within the interval [0, $10^{10}$]. The curve of frequency with boundary stiffness is shown in Fig. 13(a), where the coordinate axis scale of boundary stiffness is the logarithmic scale.

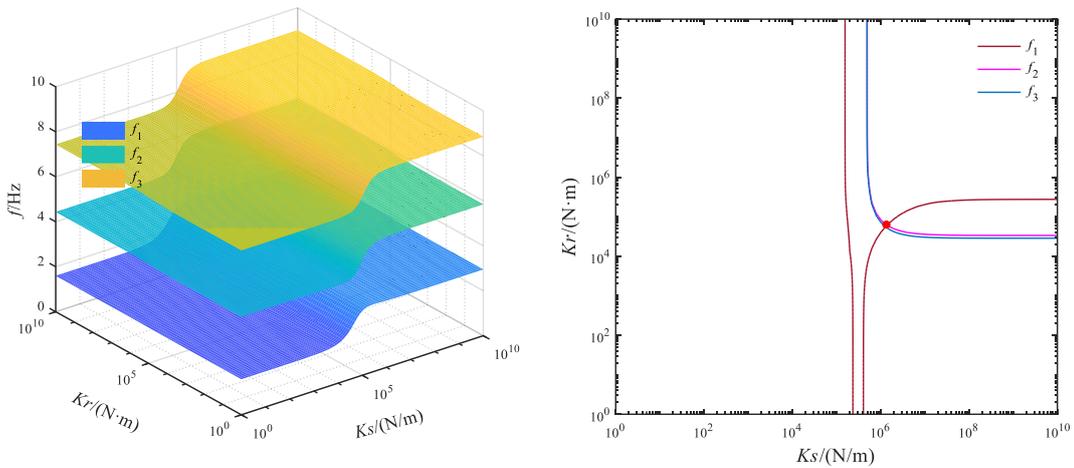

(a) Curve of multiorder frequencies of cable ① with boundary stiffness    (b) Isolines of multiorder frequencies of cable ①

Fig. 13. Parameter analysis for boundary stiffness of scenario 2 ($K_{s1} = \infty$ and $K_{r2} = \infty$).

Fig. 13(a) shows that the frequencies of different orders have the same variation trend under scenario 2. $K_r$ has a smaller impact on frequencies than $K_s$. The frequency $f$ increases with the increasing $K_r$ only under a larger $K_s$, and the region with the largest variation of frequencies with $K_s$ lay in the interval [$10^3$, $10^6$]. Similarly, according to the measured frequencies of the first three orders in the cable ①, the isolines for boundary stiffness are shown in Fig. 13(b). The three isolines approximately intersect at one point. Thus, the boundary stiffness has one and only solution under the aforementioned assumption. Comparison with scenario 1 shows that the use of the first three orders frequencies alone may not produce a unique solution of boundary stiffness in engineering practice because the lack of some constraint condition results in multiple solutions of boundary stiffness. A certain prior assumption of boundary stiffness is necessary to identify and determine the boundary stiffness value accurately.

**7.3.3 Search ranges of boundary stiffnesses of stay cables in cable-stayed bridge**

The engineering case in the present paper investigated the search ranges of the boundary stiffnesses of the suspension bridge's anchor span strands. In addition, the boundary stiffnesses of stay cables in the cable-stayed bridge are also the concern of researchers. Chen *et al.* (2016) pointed that the complicated boundary conditions of stay cables come from the anchorage devices and flexible rubber constraints at both ends of stay cables. For the range of $K_s$, Chen *et al.* (2013) identified that the $K_s$ of two representative stay cables of Chi-Lu Bridge are $5.0 \times 10^6$ N/m and $1.5 \times 10^6$ N/m, respectively, through finite element modeling analysis. Moreover, Chen *et al.* (2016) and Chen *et al.* (2018) claimed that the range of $10^4$-$10^8$ N/m basically covers the practical $K_s$ range of most stay cables. This range is also consistent with the search range of $K_s$ in the present paper. For the range of $K_r$, since there are many kinds of connections between stay cable and main girder, such as the anchor box type (e.g., Sutong Bridge) and ear plate type (or pin hinge type, e.g., Normandy Bridge), the value of $K_r$ may be any value from 0 to infinity in the actual bridge. Further, Xie and Li (2014) proposed that the cable boundary constraints may be considered rotationally rigid when $K_r > 500EI/L$.

## 8. Conclusions

The refined cable model built in the present study comprehensively considered the influence of inclination angle, sag-extensibility, flexural stiffness, and complex boundary conditions. The vibration modal equation of the cable was discretized and solved by the finite definite method. Moreover, a multiparameter identification approach was described by combining with the high-efficiency PSO algorithm. At the identification of cable tension, the proposed approach was also capable of synchronous identification of flexural stiffness, axial stiffness, boundary rotational constraint stiffness and boundary lateral support stiffness from measured multiorder frequencies. This approach exhibited excellent parameter identification performance in numerical and engineering cases, with high accuracy in cable tension identification and wide application scope.

(1) The proposed approach of cable tension identification simultaneously considered multiple cable system parameters, especially the influence of complex boundary conditions on cable vibration. This approach is greatly beneficial for improving the accuracy of cable tension identification. For numerical cases with exact values, the proposed approach can well accomplish the identification of cable tension and other parameters; moreover, the precision of cable tension identification is kept to less than 0.1%. In the engineering cases, the cable tension identified by the proposed approach has a higher accuracy than that identified using the conventional frequency-based method, frequency-based method with flexural stiffness correction, and Zui's empirical formula, suggesting that the proposed approach can adapt to the cable systems with different flexural stiffness, sag-extensibility, and boundary conditions.

(2) While ensuring accuracy, the proposed approach also can tolerate a greater initial error. Compared with the FBSU algorithm based on the Newton-Raphson method, the present approach has a lesser reliance on the choice of initial values and can perform parameter identification within the specified search range, and hence had higher applicability.

(3) The proposed approach can adapt to the engineering scenarios where the number of measured frequencies is small. It is able to identify the cable tension and other parameters based on a limited number of frequencies. The proposed approach can achieve an accurate cable tension identification using frequencies of the first three orders and higher. However, under the condition of a smaller number of measured frequencies, multiple solutions of boundary stiffness might occur, leading to the failure of boundary stiffness identification, but this phenomenon does not affect the identification of cable tension.

(4) For cables with inconspicuous sag-extensibility or large inclination angle, the static displacement perpendicular to the chordwise direction is nearly zero. As a result, the nonlinear stiffness matrix has an excessively small influence on the total stiffness matrix. In that case, the identified axial stiffness might be an impractical value. However, the present analysis shows that the mistaken identification of axial stiffness does not have a direct impact on cable tension identification. In proposed approach, the cable tension identified using the known axial stiffness is inferior to that when the axial stiffness is simultaneously identified.

(5) If the lateral support stiffness decreases, the overall frequencies of the cable decreases. If the rotational constraint stiffness is small, the cable frequencies also decreases. However, when the lateral support stiffness is small, the impact of rotational constraint stiffness on cable frequencies is not obvious. Only when the lateral support stiffness is large, the rotational constraint stiffness needs to be considered, indicating that, in addition to the flexural stiffness and sag-extensibility, the frequency-based method for cable tension measurement is also required to consider the influence of these two boundary stiffness parameters on the actual vibration frequency.

## Acknowledgments

This study was financially supported by the National Natural Science Foundation of China (Grant Nos. 52078134 and 51678148), the Natural Science Foundation of Jiangsu Province (BK20181277), the National Key R&D Program of China (No. 2017YFC0806009), and the Scientific Research Foundation of Graduate School of Southeast University (YBPY2129), which are gratefully acknowledged.